\pdfoutput=1 
\documentclass[11pt,letterpaper,logo]{preprint}
\PassOptionsToPackage{table}{xcolor}
\PassOptionsToPackage{most}{tcolorbox}

\usepackage[utf8]{inputenc}
\usepackage[T1]{fontenc}
\usepackage{microtype}

\usepackage[numbers,sort]{natbib}

\usepackage{amsmath}
\usepackage{amssymb}
\usepackage{amsfonts}
\usepackage{mathtools}
\usepackage{bm}

\usepackage{graphicx}
\usepackage{subcaption}
\usepackage{float}
\usepackage{booktabs}
\usepackage{array}
\usepackage{arydshln}
\usepackage{multirow}
\usepackage{makecell}
\usepackage{adjustbox}
\usepackage{wrapfig}
\usepackage[percent]{overpic}

\usepackage{algorithm}
\usepackage{algorithmic}

\usepackage{url}
\usepackage{xspace}
\usepackage{pifont}
\usepackage{epigraph}
\usepackage{listings}
\usepackage{enumitem}

\usepackage{etoc}

\usepackage[disable]{todonotes}

\usepackage{hyperref}
\hypersetup{hidelinks}

\usepackage{doi}
\usepackage{cleveref}

\setlist[itemize]{leftmargin=12pt}

\graphicspath{{./figures/}}

\expandafter\def\expandafter\UrlBreaks\expandafter{%
  \UrlBreaks\do\-%
}

\usepackage{amsmath,amsfonts,bm}

\def\eqref#1{equation~\ref{#1}}

\def\1{\bm{1}}

\DeclareMathAlphabet{\mathsfit}{\encodingdefault}{\sfdefault}{m}{sl}
\SetMathAlphabet{\mathsfit}{bold}{\encodingdefault}{\sfdefault}{bx}{n}

\providecommand{\hermes}{\textsc{Hermes}\xspace}
\providecommand{\learning}{\textsc{Hermes-Learn}\xspace}

\providecommand{\Lone}{\textsf{L1}\xspace}             
\providecommand{\Ltwo}{\textsf{L2}\xspace}   
\providecommand{\LtwoSS}{\textsf{L2-SS}\xspace} 
\providecommand{\LtwoSV}{\textsf{L2-SV}\xspace} 
\providecommand{\Lthree}{\textsf{L3}\xspace}    

\newtcolorbox{thmbox}{breakable, paperthm,
  left=7pt, right=7pt, top=4pt, bottom=4pt,
  before skip=6pt, after skip=6pt}
\newtcolorbox{remarkbox}{breakable, papernote,
  left=7pt, right=7pt, top=4pt, bottom=4pt,
  before skip=6pt, after skip=6pt}
\newtcolorbox{paperpromptbox}[1]{%
  breakable,
  paperprompt,
  before upper={\setlength{\parindent}{0pt}},
  left=7pt,
  right=7pt,
  top=5pt,
  bottom=5pt,
  title={#1},
  fonttitle=\small\bfseries,
  before skip=8pt,
  after skip=8pt
}

\colorlet{PaperTitleBack}{AWSViolet50}     

\definecolor{hshared}{RGB}{ 74, 85,104}   
\definecolor{hlone}{RGB}  { 59, 91,165}   
\definecolor{hltwo}{RGB}  { 47,125,110}   
\definecolor{hsub}{RGB}   {122,106, 85}   
\definecolor{hdig}{RGB}   {107, 78,140}   

\tcbset{
  lvlshared/.style={colback=hshared!5, colframe=hshared},
  lvlone/.style   ={colback=hlone!5,   colframe=hlone},
  lvltwo/.style   ={colback=hltwo!5,   colframe=hltwo},
  lvlsub/.style   ={colback=hsub!5,    colframe=hsub},
  lvldig/.style   ={colback=hdig!5,    colframe=hdig}}

\newtcolorbox[auto counter]{promptbox}[2][]{%
  breakable, enhanced, arc=1.6mm, boxrule=0.7pt,
  lvlshared, coltitle=white,
  fonttitle=\bfseries\footnotesize,
  title={Prompt~\thetcbcounter: #2},
  left=4pt, right=4pt, top=4pt, bottom=4pt,
  before skip=7pt, after skip=7pt, #1}

\newtcolorbox[auto counter]{rolloutbox}[2][]{%
  breakable, enhanced, arc=1.6mm, boxrule=0.7pt,
  lvlshared, coltitle=white,
  fonttitle=\bfseries\footnotesize,
  title={Rollout~\thetcbcounter: #2},
  left=4pt, right=4pt, top=4pt, bottom=4pt,
  before skip=7pt, after skip=7pt, #1}

\newenvironment{ptext}%
  {\par\scriptsize\ttfamily\raggedright
   \setlength{\parindent}{0pt}\setlength{\parskip}{0pt}\obeylines}%
  {\par}

\newenvironment{rtext}%
  {\par\scriptsize\raggedright
   \setlength{\parindent}{0pt}\setlength{\parskip}{0pt}\obeylines}%
  {\par}

\newcommand{\turnbar}[1]{%
  \par\addvspace{4pt}\noindent
  \colorbox{black!8}{\parbox{\dimexpr\linewidth-2\fboxsep\relax}%
    {\vspace{1pt}\sffamily\bfseries\scriptsize #1\vspace{1pt}}}%
  \par\addvspace{3pt}}

\newcommand{\boxnote}[1]{%
  \par\addvspace{3pt}{\scriptsize\itshape\textbf{Notes:} #1\par}}

\newcommand{\elide}[1]{{\scriptsize\itshape[\,#1\,]}}

\newtcolorbox{synbox}[1][]{%
  enhanced, colback=HermesBlueTint, colframe=HermesBlueTint!45!black,
  boxrule=0.7pt, arc=2mm, left=1.5mm, right=1.5mm, top=1.2mm, bottom=1.2mm,
  middle=1mm, fontupper=\scriptsize, fontlower=\scriptsize,
  before skip=0pt, after skip=0pt, #1}

\newcommand{\synA}[1]{\textcolor{HermesDeepBlue}{\textbf{#1}}}
\newcommand{\synB}[1]{\textcolor{HermesDeepGold}{\textbf{#1}}}
\newcommand{\synC}[1]{\textcolor{HermesDeepGreen}{\textbf{#1}}}

\definecolor{HermesBlue}{HTML}{335E95}
\definecolor{HermesWarmTint}{HTML}{FBF2ED}
\definecolor{HermesBlueTint}{HTML}{EDF3FA}
\definecolor{HermesBlueStrongTint}{HTML}{E3EDF8}
\definecolor{HermesReferenceTint}{HTML}{F4F4F4}

\definecolor{HermesDeepBlue}{HTML}{0F0AA1}
\definecolor{HermesDeepGold}{HTML}{D9650D}
\definecolor{HermesDeepGreen}{HTML}{1A7F37}

\lstdefinelanguage{json}{
  basicstyle=\ttfamily\footnotesize,
  stringstyle=\ttfamily,
  showstringspaces=false,
  breaklines=true,
  columns=fullflexible,
  keepspaces=true,
}

\newcommand{\cmark}{\ding{51}}
\newcommand{\xmark}{\ding{55}}

\newcommand{\std}[1]{\,{\scriptsize$\pm$\,#1}}
\newcommand{\smallstd}[1]{{\scriptsize\,\ensuremath{\pm}\,#1}}

\title{\hermes: Learning Contextual Reasoning\\
Unlocks Test-Time Scaling}

\runningtitle{HERMES: Learning Contextual Reasoning Unlocks Test-Time Scaling}

\metadata{Date}{\today}

\paperlogos{\includegraphics[height=22pt]{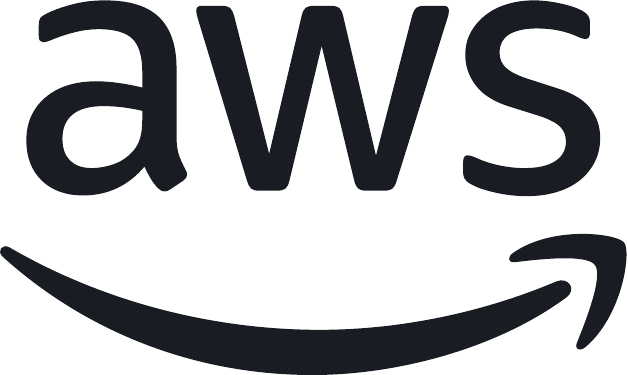}}

\hypersetup{
  pdftitle={
    HERMES: Learning Contextual Reasoning Unlocks Test-Time Scaling
  },
  pdfauthor={
    Xinyu Li,
    Mononito Goswami,
    Hao Liu,
    Nikos Kanakaris,
    Langlin Huang,
    Prithwish Jana,
    Patrick Blobaum,
    Purak Jain
  },
  pdfsubject={Machine Learning},
  pdfkeywords={
    contextual reasoning,
    test-time scaling,
    language models,
    reinforcement learning
  }
}

\begin{document}


\author{
  \normalfont
  \begin{minipage}{0.98\linewidth}
    \raggedright
    \setlength{\parskip}{0pt}

    \textbf{Xinyu Li}$^{1,\dagger,\ddagger}$,
    \textbf{Mononito Goswami}$^{2,\dagger}$,
    \textbf{Hao Liu}$^{2}$,
    \textbf{Nikos Kanakaris}$^{2}$,
    \textbf{Langlin Huang}$^{3,\ddagger}$, \\
    \textbf{Prithwish Jana}$^{4,\ddagger}$,
    \textbf{Patrick Bl\"obaum}$^{2}$,
    \textbf{Purak Jain}$^{2}$

    \par\vspace{7pt}

    $^{1}$Carnegie Mellon University
    \quad
    $^{2}$AWS AI Labs


    $^{3}$Washington University in St. Louis
    \quad
    $^{4}$Georgia Institute of Technology

  \end{minipage}
}

%
%
\begin{abstract}
Test-time scaling improves model performance by allocating additional compute during inference. Using this compute effectively across multiple context windows requires deciding how to allocate fresh contexts and what information to carry between them. We call a model's ability to make these
decisions \emph{contextual reasoning}. Existing approaches largely prescribe these decisions through their harness; we instead shift them to the model. We introduce 1) \hermes, a family of simple, configurable harnesses that progressively varies model control over context allocation and reuse, and 2) \learning, a two-stage framework for learning these capabilities. We find that capable models can exploit this flexibility to scale with additional inference-time compute, while smaller open-source models initially struggle to do so. Training with \learning closes this gap, inducing adaptive contextual reasoning strategies that vary with both the problem and the progress of reasoning. These gains generalize across benchmarks and models, extrapolate beyond the inference-time compute seen during training, and transfer to complementary test-time scaling methods beyond \hermes.
\end{abstract}

\maketitle

{\let\thefootnote\relax
\footnotetext{%
  $^{\dagger}$Xinyu Li and Mononito Goswami contributed equally. Correspondence to
  \href{mailto:xinyul2@andrew.cmu.edu}{xinyul2@andrew.cmu.edu}
  or
  \href{mailto:mononito@amazon.com}{mononito@amazon.com}.
}}

{\let\thefootnote\relax
\footnotetext{%
  $^{\ddagger}$Xinyu Li, Langlin Huang, and Prithwish Jana were interns at
  AWS when this work was carried out.
}}

\etocdepthtag.toc{mtmain}


\section{Introduction}
\label{sec:introduction}
\begin{quote}
\hfill
\begin{minipage}{0.8\textwidth}
\small\itshape
``The capacity of the human mind for formulating and solving complex problems
is very small compared with the size of the problems \ldots''

\hfill --- Herbert A. Simon, \textit{Models of Man}~\citep{simon1957models}
\end{minipage}
\end{quote}

Large language models have made impressive progress on complex reasoning, increasingly contributing to mathematical research~\citep{alpoge2026twothirds,openai2026euler}. This progress is fueled by \emph{test-time scaling}: allocating additional computation during inference to improve model performance~\citep{snell2025scaling}. Recent advances have involved thousands of agents reasoning in separate contexts, with harnesses coordinating how those contexts are allocated and how intermediate results are carried forward. This highlights two complementary dimensions of reasoning. Mathematical reasoning determines how to make progress given the information available within a context, while \emph{contextual reasoning} determines how separate contexts should be allocated, coordinated, and reused as reasoning unfolds. In this work, we focus on contextual reasoning.

Concretely, contextual reasoning requires two types of decisions: i) How should additional contexts be used in order to explore an alternative solution, verify prior work, or continue an existing line of reasoning? ii) What information should be carried forward into those contexts? 
Existing approaches typically encode these decisions in their harnesses. For example, Reasoning Cache repeatedly compresses prior reasoning and continues from the resulting summary in a fresh context~\citep{wu2026reasoning}, while Recursive Agent Optimization uses recursive delegation to create fresh contexts for subproblems and learns when and how to invoke them~\citep{gandhi2026recursive}. Yet the best use of additional contexts may change across stages of problem solving. For instance, a difficult proof may initially benefit from exploring several approaches in parallel, then verifying a promising intermediate result, and finally continuing a successful line of reasoning. This motivates a flexible harness for contextual reasoning that exposes multiple ways to use additional contexts and allows the model to choose among them based on the current state of the solution.

To this end, we introduce \hermes, a family of simple, flexible, and configurable harnesses that progressively shifts contextual reasoning control from the harness to the model. \hermes exposes two generic mechanisms: \emph{delegation}, for allocating fresh context, and \emph{digestion}, for carrying useful information forward. By varying how strongly the harness steers these decisions, \hermes defines a hierarchy ranging from prescribed strategies to increasingly flexible and model-driven contextual reasoning. We observe that capable models can exploit this flexibility, while small open-source models perform worse under \hermes than when reasoning inline, revealing a contextual reasoning gap. 
This motivates a learning framework for developing flexible contextual reasoning. Crucially, the same hierarchy provides a natural learning curriculum: models first learn under strongly steered harnesses with a simplified contextual reasoning decision space, then progressively assume greater control as harness steering is reduced during training. We therefore introduce \learning, a two-stage framework for learning flexible contextual reasoning. Together, \hermes and \learning provide a rigorous framework for studying contextual reasoning and a practical approach for teaching models to acquire this distinct capability.

We evaluate \hermes and \learning across model families and scales, from Qwen3-4B to frontier models, on mathematical reasoning benchmarks spanning AIME, HMMT, BeyondAIME~\citep{bytedance_seed_2025_beyondaime}, and IMO-AnswerBench~\citep{luong-etal-2025-towards}, as well as the out-of-domain FrontierScience benchmark~\citep{wang2026frontierscience}. We deliberately constrain each context to 8K tokens, creating a setting where effective long-horizon reasoning requires models to allocate and reuse context well. Our results show that: (1) \hermes unlocks test-time scaling for capable models, while revealing a substantial contextual reasoning gap in smaller open-source models; (2) \learning closes this gap, improving contextual reasoning without degrading the model's mathematical reasoning; (3) trained models learn adaptive contextual reasoning strategies that vary with both the problem structure and the reasoning progress; and (4) the learned contextual reasoning capabilities extend beyond the inference-time compute seen during training and transfer to complementary test-time scaling methods beyond \hermes.

\paragraph{Contributions}
\underline{(1)} \textit{Contextual reasoning.}
We study contextual reasoning as a model's capability to organize and sustain reasoning over long horizons.
\underline{(2)} \textit{\hermes.}
We introduce a simple, configurable family of harnesses built on delegation and digestion, with a steering hierarchy that progressively shifts control from the harness to the model.
\underline{(3)} \textit{A learning framework to close contextual reasoning gap.}
We show that capable models can turn additional inference-time compute into better performance under \hermes, while smaller open-source models struggle despite reasonable mathematical reasoning.
We introduce \learning, a two-stage framework combining supervised fine-tuning and curriculum reinforcement learning to close this gap by learning to select and compose delegation strategies.
\underline{(4)} \textit{Adaptive and generalizable contextual reasoning.}
Learned strategies adapt to the problem and current reasoning state, generalize across benchmarks and model families, scale beyond the inference-time compute seen during training, and transfer to complementary test-time scaling methods.

\section{Related Work}
\label{sec:related-work}
\paragraph{Contextual reasoning} Prior test-time scaling approaches typically allocate additional contexts through inference primitives such as sequential reasoning~\citep{wei2022chain, madaan2023selfrefine}, parallel exploration~\citep{cobbe2021verifiers, wang2023selfconsistency}, and aggregation~\citep{venkatraman2025rsa}, or compose multiple primitives within a single inference procedure~\citep{hu2026pacore, qin2026atlas}. Another line of work uses search to guide context allocation, such as tree-based search~\citep{yao2023tree, zhou2023language} using self-reflection and verifier-guided search~\citep{lightman2024let} using process reward models. When reasoning extends across multiple context windows, existing work primarily relies on delegation and compaction to coordinate context allocation and reuse. Prior methods typically prescribe specific delegation strategies, such as delegating recursively decomposed tasks~\citep{prasad2024adapt}, recursively spawned work~\citep{schroeder2024thread}, or concurrently executable structures~\citep{chi2025asyncthink} to fresh contexts. A complementary line of work~\citep{zhou2025mem1,kontonis2026memento} focuses on compressing prior reasoning into compact states for reuse in later contexts.

While prior work relies on external search procedures or prescribes contextual reasoning strategies in the inference procedure, \hermes studies contextual reasoning as a model capability. \hermes enables models to adaptively choose among multiple strategies for allocating additional contexts as reasoning progresses. Moreover, our experiments (Sec.~\ref{sec:exp:generalization}) show that this capability can also benefit complementary test-time scaling methods that use search with self-reflection.
 
\begin{wraptable}[21]{r}{0.55\textwidth}
\centering
\small
\caption{
Comparison of \hermes with existing methods. 
Prior methods mostly prescribe contextual reasoning strategies for allocating and reusing contexts, then train models for that particular harness. \learning instead trains models to select among different contextual reasoning strategies through the learning curriculum induced by the varying steering of the \hermes hierarchy.
\textit{Model-driven strategy selection} captures whether contextual reasoning strategies are decided by the model rather than by the harness;
\textit{Context compaction}, whether prior reasoning can be compressed and reused; 
\textit{Recursive delegation}, whether further contexts can be allocated recursively through the same delegation interface.
}
\label{tab:related-work-comparison}

\resizebox{\linewidth}{!}{%
\begin{tabular}{@{}rcccl@{}}
\toprule
Method
& \makecell{Model-driven\\strategy selection}
& \makecell{Context\\compaction}
& \makecell{Recursive\\delegation} \\
\midrule

\texttt{SPIRAL}
& \xmark
& \xmark
& \xmark \\

\texttt{PaCoRe}
& \xmark
& \cmark
& \xmark \\

\texttt{RAO}
& \xmark
& \xmark
& \cmark \\

\texttt{Context-Folding}
& \xmark
& \cmark$^\dagger$
& \xmark \\

\texttt{Reasoning Cache}
& \xmark
& \cmark
& \xmark \\

\midrule
\textbf{\hermes}
& \cmark
& \cmark
& \cmark \\

\bottomrule
\end{tabular}%
}

\vspace{2pt}
\scriptsize
$^\dagger$Context-Folding compacts a delegated branch when it returns.
\end{wraptable}

\paragraph{Learning contextual reasoning}
Recent work shows that explicitly training models to use the test-time inference procedures can improve test-time scaling. For example, \texttt{MRT} shapes long reasoning trajectories with progress-sensitive rewards \citep{qu2025mrt}, and \textsc{e3} trains models to compose generation, verification, and refinement during exploration \citep{setlur2025e3}. Other methods train models for particular inference harnesses: \texttt{SPIRAL} trains parallel generation and aggregation \citep{hamid2026spiral}, \texttt{PaCoRe} trains synthesis across parallel branches~\citep{hu2026pacore}, \texttt{RAO} trains recursive delegation \citep{gandhi2026recursive}, \texttt{Context-Folding} trains delegation of subtasks to fresh branches together with compaction of intermediate reasoning~\citep{sun2025contextfolding}, and \texttt{Reasoning Cache} trains continuation from compacted reasoning progress~\citep{wu2026reasoning}. While these methods demonstrate the value of aligning training with inference harnesses, they typically encode contextual reasoning strategies in the harness, leaving models to learn how to execute them effectively (App.~\ref{app:related-work}). In contrast, \learning teaches models to autonomously select, instantiate, and compose different contextual reasoning strategies, through the natural curriculum induced by the progressively reduced steering of the \hermes hierarchy.

\section{\hermes: A Family of Simple, Flexible Harnesses}
\label{sec:hermes}

\begin{wrapfigure}[19]{l}{0.55\textwidth}
\centering
\includegraphics[width=\linewidth, trim=15 10 0 22, clip]{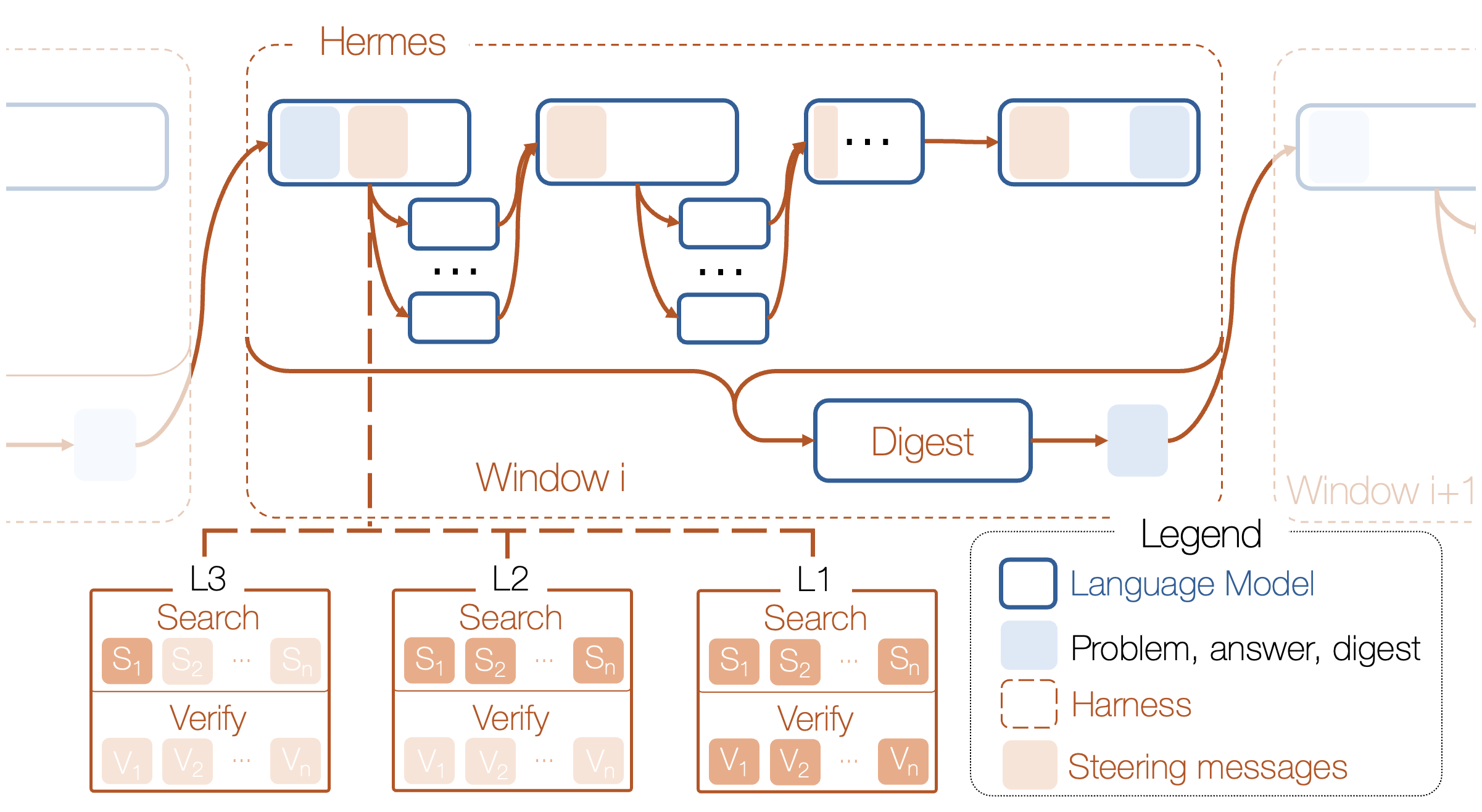}
\caption{
\textbf{Overview of \hermes.} \hermes enables contextual reasoning across multiple context windows through two mechanisms: \emph{delegation}, which allocates fresh contexts to search or verification strategies, and \emph{digestion}, which compacts accumulated progress for reuse in later contexts. The \hermes hierarchy (\Lthree--\Lone) progressively shifts control over delegation decisions from the harness to the model. \Lthree prescribes the strategy; \Ltwo prescribes a strategy category (search or verification) while allowing the model to select strategies within it; and \Lone lets the model choose both as reasoning unfolds.
}
\label{fig:hermes_overview}
\end{wrapfigure}

In this section, we introduce \hermes and evaluate how different language models can leverage its harnesses for test-time scaling. 

\paragraph{Desiderata} We design \hermes to be \emph{simple}, \emph{flexible}, and \emph{configurable}.
\emph{Simple}: \hermes introduces only two generic mechanisms, \emph{delegation} and \emph{digestion} to allocate fresh contexts and carry useful information across them. 
\emph{Flexible}: these mechanisms do not prescribe a fixed form of computation; instead, they allow the model to decide how additional contexts should be used as reasoning unfolds. 
\emph{Configurable}: steering controls how much guidance the harness provides for delegation, from prescribing a specific strategy to letting the model choose one itself.

\paragraph{Contextual reasoning via delegation and digestion}
\label{sec:hermes:delegation_digesting}
\hermes gives a model two ways to continue reasoning beyond a single context window. It can assign tasks to fresh contexts and use the results, or carry its own progress into a fresh context and continue working. We call these operations \emph{delegation} and \emph{digestion}, respectively. Delegation and digestion are complementary primitives for contextual reasoning, expanding reasoning across contexts in parallel and sequentially (Fig.~\ref{fig:hermes_overview}).

Within each context window, \hermes runs a ReAct-style interaction loop~\citep{yao2023react} (Fig.~\ref{fig:hermes_overview}). At step $t$, the agent receives an input $u_t$, such as the original problem, a message from the harness, or a result from a subagent. It then produces an action $a_t$: it may continue reasoning, return an answer, or call \texttt{launch\_subagent(task)}. This tool is how the agent delegates work. The harness invokes the same model on the specified task in a fresh context and returns a compact report as the agent's next input. We call this model working on the specified task a \emph{subagent}. Formally, $h_t=(u_1,a_1,\ldots,u_t)$ is the history available when the agent chooses $a_t$; under harness \(H\) with configuration $c$, the action is sampled from $\pi_\theta(\cdot\mid h_t;H_c)$.
An agent has up to $T$ steps in each context window. At each of the first $T-1$ steps, it may launch up to $B$ subagents; at step $T$, it can no longer delegate and returns its best current answer. Subagents use the same delegation interface, so they may launch further subagents up to depth $D$. At the maximum depth, they reason inline without further delegation. The interaction can also end earlier if the context window is exhausted.

Digestion allows the agent to continue \emph{its own} work across windows. At the end of a window, it compresses its accumulated progress into a digest. If reasoning continues, the harness starts a fresh window with the original problem and that digest, rather than the full interaction history, for up to $W$ windows. Thus, $B$ and $D$ govern how work branches across contexts in parallel, while $W$ governs how long it can continue sequentially.

\paragraph{Steerable control over delegation}
\label{sec:hermes:hierarchy}
Effective delegation requires more than sound mathematical reasoning. An agent must also decide when to delegate, what task to assign, and how to use each fresh context, for example by independently searching for a solution, exploring a different approach, decomposing the problem, or verifying existing reasoning. These contextual reasoning decisions can be challenging, particularly for smaller open-source models. \hermes therefore varies how strongly the harness steers delegation, progressively shifting these decisions from the harness to the model. We organize delegation strategies into two broad categories: \emph{search} through independent attempts on the full problem, alternative approaches, or decomposition into simpler subproblems, and \emph{verification} of complete solutions, specific reasoning steps, or disagreements between candidate solutions. These categories induce three steering levels $\ell \in$ \{\Lthree, \Ltwo, \Lone\} (Fig.~\ref{fig:hermes_overview}). At \Lthree, the harness specifies a strategy and the agent instantiates it as a concrete task. At \Ltwo, the harness specifies only the strategy category, while the agent selects one or more strategies within it. At \Lone, the agent selects from the full strategy set, deciding how to formulate tasks for its subagents as reasoning unfolds. Thus, \Lthree--\Lone progressively shift control over delegation from the harness to the agent, while providing a natural curriculum for learning these contextual reasoning decisions. Several prior methods can be viewed as fixed-strategy special cases of \hermes~\citep{wu2026reasoning,gandhi2026recursive,hamid2026spiral}.

Together, these parameters define a harness configuration $c=(T,B,D,W,\ell)$, making \hermes configurable in both how much additional context is available and how much control over its use is given to the model. Additional details can be found in App.~\ref{app:harness}.

\paragraph{Test-time scaling with \hermes relies on contextual reasoning} \hermes provides additional test-time compute, but models must use it effectively to improve performance. We compare \texttt{Qwen3-4B-\allowbreak{}Instruct-\allowbreak{}2507} (hereafter \texttt{Qwen3-4B}), our base model, with Claude \texttt{Sonnet\allowbreak{}5} and \texttt{DeepSeek-\allowbreak{}V3.2} under \emph{inline reasoning} and \hermes \Lone and \Ltwo. Inline reasoning measures mathematical 
\begin{wraptable}[15]{r}{0.55\textwidth}
\centering
\caption{\textit{\hermes enables test-time scaling for capable models.} $\mathrm{Avg}@16$ (\%, $\uparrow$) under inline reasoning and \hermes \Lone and \Ltwo on the pooled and synthetic benchmarks. Inline reasoning measures mathematical reasoning within a single 8K-token context, while \hermes provides up to 56K tokens across contexts. Claude \texttt{Sonnet 5} and \texttt{DeepSeek-V3.2} match or improve over inline reasoning with \hermes; \texttt{Qwen3-4B} degrades, exposing a contextual reasoning gap.} 
\label{tab:inline-vs-harness}
\setlength{\tabcolsep}{4pt}
\resizebox{\linewidth}{!}{%
\begin{tabular}{@{}lccc@{\hspace{1em}}ccc@{}}
\toprule
& \multicolumn{3}{c}{Pooled ($n=238$)}
& \multicolumn{3}{c}{Synthetic ($n=88$)} \\
\cmidrule(lr){2-4}
\cmidrule(l){5-7}
Model & Inline & \Lone & \Ltwo
      & Inline & \Lone & \Ltwo \\
\midrule
\rowcolor{HermesWarmTint}
\texttt{Qwen3-4B}
& \textbf{37.1} & 16.2 & 15.1
& \textbf{50.7} & 26.8 & 26.0 \\
\midrule
\texttt{Sonnet 5}
& 55.3 & \textcolor{HermesBlue}{\textbf{76.7}} & 76.5
& 87.0 & 94.7 & \textcolor{HermesBlue}{\textbf{96.2}} \\
\texttt{DeepSeek-V3.2}
& 56.2 & \textcolor{HermesBlue}{\textbf{58.4}} & 57.4
& 78.1 & \textcolor{HermesBlue}{\textbf{86.2}} & 83.7 \\
\bottomrule
\end{tabular}%
}
\end{wraptable}
reasoning within a single 8K-token context, where a non-thinking model solves the problem directly without delegation or digestion. For both \hermes harnesses, we use $T=3$, $B=3$, $D=1$, and $W=1$, with an 8K-token context per agent. We set \(D=1\) and \(W=1\), so only the root agent delegates and digestion 
is disabled. With \(T=3\) and \(B=3\), \hermes can use up to 56K tokens\footnote{With $T=3$, $B=3$, $D=1$, and $W=1$, the root agent uses one 8K-token window and may launch up to six subagents, each with its own 8K-token window. This yields a maximum of $7\times 8\mathrm{K}=56\mathrm{K}$ tokens.} of total context, compared with 8K tokens for inline reasoning. We evaluate whether models can translate this additional context into improved performance on a pooled set of AIME and HMMT problems (App.~\ref{app:exp:pooled-benchmark}) and a synthetic benchmark constructed to benefit from different contextual reasoning strategies (App.~\ref{app:aug:synthetic-construction}).

As shown in Tab.~\ref{tab:inline-vs-harness}, Claude \texttt{Sonnet 5} and \texttt{DeepSeek-V3.2} match or exceed their inline performance with \hermes, showing that they can combine additional context with mathematical reasoning effectively. In contrast, \texttt{Qwen3-4B} degrades substantially despite reasonable inline mathematical reasoning, revealing a gap in the contextual reasoning needed to benefit from additional context. On the synthetic benchmark, \texttt{Sonnet 5} also performs slightly better under the more prescriptive \Ltwo harness than under \Lone, suggesting that even capable models may benefit from different levels of steering. These results motivate learning contextual reasoning capabilities directly.

\section{Learning Contextual Reasoning}
\label{sec:learning}

\hermes exposes several learning problems in contextual reasoning. Given the current history $h_t$, a model must learn \emph{what to delegate} to fresh contexts, \emph{what to preserve} when generating a digest, and \emph{how to reason} from the resulting digest. Prior work has studied the latter two problems: models can be trained to continue reasoning from compressed histories~\citep{wu2026reasoning, wu2026resum}, while recent instruction-tuned models already exhibit summarization capabilities that enable effective digest generation in many cases. Further improving digestion through downstream task outcomes can suffer from difficult credit assignment across multiple subsequent context windows.

In this work, we focus on learning delegation. Unlike digestion, delegation decisions are made by the root agent as reasoning unfolds and can be optimized directly from task outcomes. We introduce \learning, a two-stage framework that first teaches useful delegation behaviors through supervised fine-tuning (SFT), then develops autonomous strategy selection through reinforcement learning (RL). For digestion, we rely on the model's existing ability to summarize its accumulated reasoning and continue from the resulting digest instead of explicitly training it.

\paragraph{Stage I: Supervised Fine-Tuning}
We first warm-start the base model with supervised fine-tuning on successful teacher trajectories. We collect trajectories from strong models operating under \Lone and \Ltwo, exposing the model to delegation under different levels of steering. We retain trajectories in which the root agent delegates at least once and produces the correct final answer. During training, since we set \(D=1\), all delegation decisions occur at the root, so we train only on root-agent trajectories and retain subagent outputs only as part of the root history \(h_t\). Let $\mathcal{T}$ denote these retained root-agent trajectories. Then the SFT stage optimizes
\begin{equation*} 
\arg\min_{\theta}
-\sum_{\tau\in\mathcal{T}}
\sum_{t=1}^{T}
\sum_{j=1}^{|a_t|}
\log
\pi_\theta
\left(
a_{t,j}
\mid
h_t,a_{t,<j}
\right),
\end{equation*}
where loss is computed only over assistant tokens. This stage exposes the model to successful delegation under different levels of steering and provides a strong initialization for RL.

\paragraph{Stage II: Reinforcement learning}
Initializing from the SFT policy \(\pi_{\theta_{\mathrm{SFT}}}\), we use on-policy reinforcement learning with outcome-based rewards to learn more autonomous delegation. We use Dr.~GRPO~\citep{liu2025understanding} with the DAPO clip-higher strategy and dynamic sampling~\citep{yu2025dapo}. For each problem \(p \in \mathcal{D}\) and \hermes harness \(H\), we sample \(G\) root-agent rollouts from the current policy. Each rollout receives a binary reward of \(1\) if its final answer is correct and \(0\) otherwise. Following dynamic sampling~\citep{yu2025dapo}, we train only on groups containing both correct and incorrect rollouts.

As in Stage I, we optimize only the root-agent trajectory. Extending training to subagent trajectories would additionally require optimizing reasoning over policy-generated subtasks and assigning credit to individual subagent outcomes. Recursive Agent Optimization (RAO)~\citep{gandhi2026recursive}, for example, addresses this setting by jointly optimizing nodes in a recursive execution tree using node-local rewards and trajectory weighting. In this work, we focus on learning root-level delegation decisions and leave joint root--subagent optimization to future work.

We optimize root-agent trajectories with the following Dr.~GRPO objective:
\begin{equation*} 
\resizebox{\linewidth}{!}{$
\begin{aligned}
\mathcal{J}(\pi_\theta) 
&= 
\mathbb{E}_{
p \sim \mathcal{D},
H \sim S(\mathcal{H}),
\{o_i\}_{i=1}^{G} \sim \pi_{\theta_{\mathrm{old}}}(\cdot \mid p; H)}
\Bigg[ 
\frac{1}{G} 
\sum_{i=1}^G 
\sum_{t=1}^{T} 
\sum_{j=1}^{|a^i_t|} 
\min 
\Big( 
r^i_{t,j}(\pi_\theta) \hat A_i,
\,
\mathrm{clip}\!\left(
r^i_{t,j}(\pi_\theta), 
1 - \epsilon_{\mathrm{low}}, 
1 + \epsilon_{\mathrm{high}} 
\right) 
\hat A_i 
\Big) 
\Bigg]
\end{aligned},
$}
\label{eq:grpo}
\end{equation*}

where
$$
r_{t,j}^i(\pi_\theta)
=
\frac{
\pi_\theta(a_{t,j}^i \mid h_t^i,a_{t,<j}^i)
}{
\pi_{\theta_{\mathrm{old}}}(a_{t,j}^i \mid h_t^i,a_{t,<j}^i)
},
\qquad
\hat{A}_i
=
R_i-\frac{1}{G}\sum_{k=1}^{G}R_k.
$$

Here, \(R_i\) is the binary outcome reward for rollout \(i\). Since the reward is defined at the rollout level, the same advantage \(\hat A_i\) is applied to all root-agent tokens in that rollout.

The sampler \(S(\mathcal{H})\) defines a curriculum over the \hermes hierarchy. We begin RL with \Ltwo and gradually increase the proportion of \Lone, while retaining a small fraction of \Ltwo throughout training. This progressively reduces harness steering, requiring the model to take greater control over which delegation strategies to use as reasoning unfolds.

\begin{wrapfigure}[16]{r}{0.5\textwidth}
\vspace{-0.6\baselineskip}
\begin{tcolorbox}[
    colback=HermesBlueTint,
    colframe=HermesBlueTint!45!black,
    boxrule=0.7pt,
    arc=2mm,
    left=2mm,
    right=2mm,
    top=2mm,
    bottom=2mm,
]
\small
Let \(n\) be a natural number, and let \(f(n)\) denote the sum of the digits
of \(n^2+1\) when written in base \(10\). Define the iterates
\(f^{(1)}(n)=f(n)\) and \(f^{(k+1)}(n)=f(f^{(k)}(n))\) for \(k\geq1\).
Let \(\textcolor{HermesDeepBlue}{\textbf{m}}\) be the value of \(f^{(100)}(1990)\).

\smallskip

Let \(\textcolor{HermesDeepGold}{\textbf{p}}\) be the number of four-digit numbers whose digits sum to \(12\).

\smallskip

Find the remainder when \(\textcolor{HermesDeepBlue}{\textbf{m}} + \textcolor{HermesDeepGold}{\textbf{p}}\) is divided by \(1000\).

\tcblower

\footnotesize
\textbf{Golden answer:} \(\boxed{353}\)
\end{tcolorbox}
\caption{\textit{Synthetic splitting example.} Constructed by combining two independent \texttt{AceReason} questions into a single problem whose final answer depends on both, encouraging parallel delegation. }
\label{fig:synthetic-splitting}

\end{wrapfigure}

\paragraph{Data augmentation} Competition math problems often do not expose delegation strategies uniformly: for example, independent attempts may arise more frequently than decomposition. To provide more balanced coverage of delegation strategies during training, we generate synthetic problems with structures designed to favor different search and verification strategies. \emph{Splitting} combines multiple independent problems into a single instance whose final answer depends on all of them, creating opportunities for parallel delegation (Fig.~\ref{fig:synthetic-splitting}). \emph{Chaining} introduces dependencies that favor sequential delegation; \emph{verifying} introduces potentially incorrect intermediate results; and \emph{multi-} and \emph{single-attempt} problems vary whether additional independent attempts are useful (see details in App.~\ref{app:aug}).

These controlled structures expose the model to a broader and more balanced range of contextual reasoning behaviors and encourage it to adapt its delegation strategy to the problem. We use the same generation procedure to construct both synthetic training examples and a held-out synthetic evaluation set.

\section{Experimental Setup and Main Results}
\label{sec:experiments}

We focus on answering \underline{four} research questions: \textbf{\textcolor{HermesBlue}{(RQ1)}} Does \learning improve contextual reasoning? \textbf{\textcolor{HermesBlue}{(RQ2)}} What contextual reasoning behaviors emerge through training? \textbf{\textcolor{HermesBlue}{(RQ3)}} Which components of the training recipe drive these improvements? \textbf{\textcolor{HermesBlue}{(RQ4)}} Do these gains generalize to new settings? We first describe the training and evaluation setup and baselines, and then address each question in turn.

\paragraph{Training setup} We train the \texttt{Qwen3-4B-Instruct-2507} base model on \textsc{AceReason-5K}, a cleaned subset of AceReason-Math~\citep{chen2026acereason} containing 5,239 mathematical reasoning problems.  For SFT, we train on teacher trajectories collected from both \texttt{Sonnet 5} and \texttt{DeepSeek-V3.2}, using \Lone and \Ltwo harnesses.
For RL, we use a curriculum that shifts from \Ltwo to \Lone while retaining balanced exposure to both harnesses throughout training, and augment training set with 500 synthetic problems. See App.~\ref{app:exp:training} for details.

\paragraph{Evaluation benchmarks and protocols} 
\label{sec:exp:evaluation}
We evaluate on a suite of mathematical reasoning benchmarks: \underline{(1)} a cleaned \textbf{pooled} benchmark of 238 AIME and HMMT problems from competitions between 2023 and 2026; \underline{(2)} a \textbf{synthetic} benchmark of 88 problems generated from the pooled benchmark using the same augmentation pipeline as our training data; and \underline{(3)} advanced benchmarks, BeyondAIME~\citep{bytedance_seed_2025_beyondaime} and IMO-AnswerBench~\citep{luong-etal-2025-towards}, together with the out-of-domain FrontierScience benchmark~\citep{wang2026frontierscience}. We use an 8K-token context per agent, a practical and deliberately constrained setting that makes reasoning across multiple contexts necessary. We report average accuracy over 16 sampled solutions ($\mathrm{Avg}@16$) under both inline reasoning and \hermes \Lone. Inline performance reflects mathematical reasoning within a single context, while improvements under \Lone reflect the model's ability to use additional contexts effectively. Training and evaluation use the default \hermes configuration: $T=3$, $B=3$, $D=1$, and $W=1$. Under this configuration, only the root agent delegates, with up to three subagents per delegation step, and digestion is disabled (see App.~\ref{app:exp}).

\paragraph{Baselines} In addition to the base model, we include two baselines designed to isolate the source of improvements from \learning. First, we evaluate inline reasoning with a $56$K-token context, matching the maximum total context available to a 3-step \hermes run\footnote{With $T=3$ and $B=3$, \hermes uses one root context and up to six subagent contexts, for a maximum of $7\times8\mathrm{K}=56\mathrm{K}$ tokens. We include this baseline to distinguish gains from reasoning across contexts from those obtained by simply providing a larger single context.}. Second, we train with GRPO under inline reasoning with an $8$K-token context, controlling for improvements from RL without training the model to reason across contexts.

\begin{table}[!bt]
\centering
\caption{SFT and RL progressively improve \Lone performance ($\mathrm{Avg}@16$ (\%, $\uparrow$)), while inline performance changes relatively little. SFT + RL under \Lone also exceeds the base model's 56K inline reference on all four mathematical benchmarks. Trained results are mean $\pm$ standard deviation over three seeds. \textbf{Bold} and \underline{underlined} mark the best and second-best point estimates per dataset.}
\label{tab:hermes-learn}

\small
\setlength{\tabcolsep}{2.5pt}
\renewcommand{\arraystretch}{1.15}
\begin{adjustbox}{max width=\linewidth}
\begin{tabular}{@{}l*{5}{@{\hspace{0.7em}}cc}@{}}
\toprule
& \multicolumn{2}{c}{Pooled}
& \multicolumn{2}{c}{Synthetic}
& \multicolumn{2}{c}{BeyondAIME}
& \multicolumn{2}{c}{IMO-Answer}
& \multicolumn{2}{c}{FrontierScience} \\
\cmidrule(lr){2-3}
\cmidrule(lr){4-5}
\cmidrule(lr){6-7}
\cmidrule(lr){8-9}
\cmidrule(l){10-11}
Model & Inline & \Lone & Inline & \Lone & Inline & \Lone
      & Inline & \Lone & Inline & \Lone \\
\midrule
Base
  & 37.1 & \cellcolor{HermesWarmTint}16.2
  & 50.7 & \cellcolor{HermesWarmTint}26.8
  & 24.8 & \cellcolor{HermesWarmTint}12.2
  & 36.2 & \cellcolor{HermesWarmTint}18.2
  & 23.6 & \cellcolor{HermesWarmTint}8.8 \\
GRPO
  & 40.4\smallstd{1.0}
  & \cellcolor{HermesWarmTint}16.5\smallstd{1.6}
  & 61.5\smallstd{1.6}
  & \cellcolor{HermesWarmTint}27.9\smallstd{1.5}
  & \underline{31.1}\smallstd{0.6}
  & \cellcolor{HermesWarmTint}14.1\smallstd{1.4}
  & 37.4\smallstd{0.6}
  & \cellcolor{HermesWarmTint}19.0\smallstd{1.4}
  & \underline{24.6}\smallstd{1.5}
  & \cellcolor{HermesWarmTint}8.9\smallstd{0.6} \\
\addlinespace[0.4ex]
Base (56K)
  & \cellcolor{HermesReferenceTint}\underline{41.8} & --
  & \cellcolor{HermesReferenceTint}59.7 & --
  & \cellcolor{HermesReferenceTint}\underline{31.1} & --
  & \cellcolor{HermesReferenceTint}\underline{40.3} & --
  & \cellcolor{HermesReferenceTint}\underline{24.6} & -- \\
GRPO (56K)
  & \cellcolor{HermesReferenceTint}41.3\smallstd{1.2} & --
  & \cellcolor{HermesReferenceTint}\underline{62.2}\smallstd{2.1} & --
  & \cellcolor{HermesReferenceTint}30.9\smallstd{1.2} & --
  & \cellcolor{HermesReferenceTint}37.5\smallstd{0.6} & --
  & \cellcolor{HermesReferenceTint}\textbf{25.1}\smallstd{1.5} & -- \\
\midrule
\multicolumn{11}{@{}l}{\textcolor{HermesBlue}{\textbf{\learning}}} \\
SFT
  & 35.3\smallstd{0.3}
  & \cellcolor{HermesBlueTint}39.9\smallstd{0.7}
  & 52.7\smallstd{1.6}
  & \cellcolor{HermesBlueTint}57.2\smallstd{2.4}
  & 23.5\smallstd{0.4}
  & \cellcolor{HermesBlueTint}31.0\smallstd{0.6}
  & 36.7\smallstd{0.5}
  & \cellcolor{HermesBlueTint}39.2\smallstd{0.2}
  & 18.4\smallstd{1.4}
  & \cellcolor{HermesBlueTint}18.9\smallstd{0.5} \\
SFT + RL
  & 36.4\smallstd{0.1}
  & \cellcolor{HermesBlueTint}\textbf{47.3}\smallstd{1.3}
  & 55.2\smallstd{0.9}
  & \cellcolor{HermesBlueTint}\textbf{74.0}\smallstd{2.9}
  & 27.5\smallstd{1.3}
  & \cellcolor{HermesBlueTint}\textbf{38.5}\smallstd{0.5}
  & 36.4\smallstd{0.4}
  & \cellcolor{HermesBlueTint}\textbf{44.0}\smallstd{0.8}
  & 19.1\smallstd{0.4}
  & \cellcolor{HermesBlueTint}24.1\smallstd{0.9} \\
\bottomrule
\vspace{-10pt}
\end{tabular}
\end{adjustbox}
\end{table}

\paragraph{\textcolor{HermesBlue}{(RQ1)} \learning enables test-time scaling with \hermes} It transforms additional context from a liability into a source of test-time scaling (Tab.~\ref{tab:hermes-learn}). The base model performs substantially worse under \Lone than inline reasoning across all benchmarks, showing that access to additional contexts alone is insufficient. SFT reverses this trend, and RL further improves \Lone performance. In contrast, inline performance changes relatively little through \learning, indicating that the gains primarily come from learning to use additional contexts rather than from improved mathematical reasoning. This distinction is reinforced by the inline GRPO baseline, which improves inline performance but continues to degrade under \Lone. Finally, SFT + RL under \Lone outperforms the 56K inline reference of both the base and GRPO models across all four mathematical benchmarks, showing that improved mathematical reasoning alone is insufficient to realize the benefits of additional context and highlighting the role of \learning in enabling effective contextual reasoning.

\begin{wrapfigure}[13]{r}{0.53\linewidth}
    \centering
    \includegraphics[width=\linewidth]{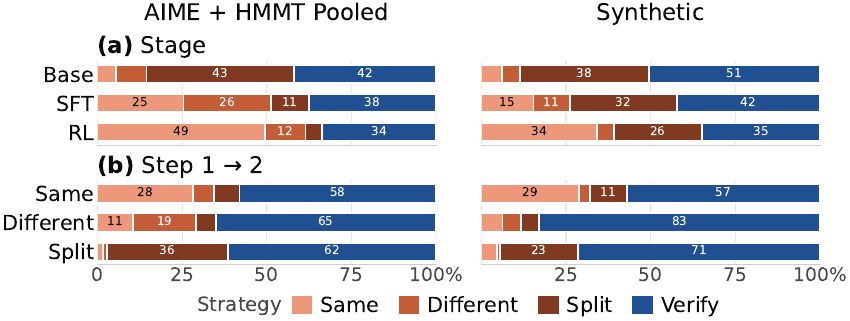}
    \caption{\learning makes strategy selection problem- and progress-dependent. \texttt{Split} is used more often on synthetic problems designed to benefit from decomposition, while the second delegation step increasingly favors verification or continues the first-step strategy. 
    }
    \label{fig:strategy-selection}
\end{wrapfigure}

\paragraph{\textcolor{HermesBlue}{(RQ2)} Training makes strategy selection \mbox{problem} and progress-dependent}
\label{sec:exp:behavior}
To understand how training changes the use of \hermes, we use an LLM judge to label the problem solving strategies executed under \Lone on the pooled and synthetic benchmarks (App.~\ref{app:exp:judge}).  Fig.~\ref{fig:strategy-selection} reveals two patterns. First, strategy selection depends on the problem: \texttt{Split} becomes much less common on the pooled benchmark after training, but remains prevalent on synthetic problems designed to benefit from decomposition (see App.~\ref{app:harness:hierarchy} for strategy definitions). Second, strategy selection depends on reasoning progress: in the second delegation step, the model often switches to verification; otherwise, it tends to continue the strategy used in the first step. Together, these results show that \learning does not learn a fixed delegation pattern, but selects strategies based on the problem and the progress made so far. Additional analyses and example rollouts are provided in Apps.~\ref{app:exp:behavior} and~\ref{app:rollouts}.

\section{Analysis and Discussion}
\label{sec:exp:learning}
\subsection{\textbf{\textcolor{HermesBlue}{(RQ3)}} Teacher diversity, curriculum, and synthetic data shape \learning.}
\label{sec:exp:ablation}
We ablate the main components of \learning on the pooled benchmark (Tab.~\ref{tab:hermes-ablations}). For SFT, we vary the source and amount of teacher demonstrations. For RL, we study exposure to \Lone and \Ltwo harnesses, their training order, and data augmentation.

\textbf{Combining teachers yields the strongest SFT model.}
Training on the full set of trajectories from both \texttt{Sonnet 5} and \texttt{DeepSeek-V3.2} gives the best SFT performance (\underline{Both (Full)}). When we match the mixed-teacher dataset to the token budget of a single teacher (\underline{Both (Matched)}), however, its performance is similar to \texttt{DeepSeek-V3.2} alone. This suggests that much of the gain from the full dataset comes from having more high-quality demonstrations rather than teacher diversity alone. Notably, \texttt{DeepSeek-V3.2} is a much better teacher than \texttt{Sonnet 5}, despite performing worse under \hermes (Tab.~\ref{tab:inline-vs-harness}), indicating that stronger inference performance does not necessarily translate into better supervision for SFT.

\begin{table*}[t]
\small
\centering

\caption{
\textbf{\learning ablations} on the pooled benchmark.
\textbf{(a)} For SFT, \textit{Both (Matched)} uses a token-matched subset
of trajectories from both teachers, while \textit{Both (Full)} uses the
full combined dataset.
\textbf{(b)} For RL, \textit{Mix} interleaves \Lone and \Ltwo, while
\textit{Curriculum} trains on \Ltwo before \Lone with matched exposure;
\textit{Rephrased} and \textit{Synthetic} each add 500 problems.
$\mathrm{Avg}@16$ (\%, $\uparrow$), mean $\pm$ s.d. over three seeds.
\textbf{Bold} and \underline{underlined} mark best and second-best results.
}
\label{tab:hermes-ablations}

\vspace{0.4em}

\setlength{\tabcolsep}{5pt}
\renewcommand{\arraystretch}{1.08}

\begin{minipage}[t]{0.36\textwidth}
\vspace{0pt}
\centering

\textbf{(a) SFT ablations}
\par\vspace{0.35em}

\begin{adjustbox}{max width=\linewidth}
\begin{tabular}{@{}lcc@{}}
\toprule
Teachers & Inline & \Lone \\
\midrule

\texttt{Sonnet}
    & 31.5\std{0.5}
    & \cellcolor{HermesBlueTint}30.1\std{0.9} \\

\texttt{DeepSeek}
    & \underline{34.5}\std{0.4}
    & \cellcolor{HermesBlueTint}\underline{39.3}\std{0.2} \\

Both (Matched)
    & 34.4\std{0.7}
    & \cellcolor{HermesBlueTint}38.9\std{0.4} \\

\midrule

Both (Full)
    & \textbf{35.3}\std{0.3}
    & \cellcolor{HermesBlueTint}\textbf{39.9}\std{0.7} \\

\bottomrule
\end{tabular}
\end{adjustbox}

\end{minipage}
\hfill
\begin{minipage}[t]{0.61\textwidth}
\vspace{0pt}
\centering

\textbf{(b) RL ablations}
\par\vspace{0.35em}

\begin{adjustbox}{max width=\linewidth}
\begin{tabular}{@{}lccc@{}}
\toprule
Schedule & Data Aug. & Inline & \Lone \\
\midrule

\multicolumn{4}{@{}l}{\emph{Harness schedule on original data}} \\

\Lone{} only
    & --
    & 36.5\std{1.5}
    & 46.1\std{0.8} \\

\Ltwo{} only
    & --
    & 36.2\std{0.2}
    & 45.4\std{0.6} \\

\addlinespace[0.3ex]

Mix
    & --
    & 35.7\std{0.7}
    & 46.1\std{0.8} \\

Curriculum
    & --
    & 36.3\std{0.7}
    & \underline{46.3}\std{1.9} \\

\midrule

\multicolumn{4}{@{}l}{\emph{With data augmentation}} \\

Mix
    & + Synthetic
    & 35.9\std{1.0}
    & 45.3\std{1.8} \\

\rowcolor{HermesBlueTint}
Curriculum
    & + Synthetic
    & 36.4\std{0.1}
    & \textbf{47.3}\std{1.3} \\

Curriculum
    & + Rephrased
    & 36.0\std{0.3}
    & 45.6\std{1.2} \\

\bottomrule
\end{tabular}
\end{adjustbox}

\end{minipage}

\end{table*}

\textbf{Curriculum provides the largest gains when paired with synthetic problems.} We ablate two choices in RL training: how the model is exposed to \Lone and \Ltwo, and exposure to synethic data. Mix and Curriculum use the same number of \Lone and \Ltwo episodes, differing only in their order. Similarly, we compare 500 synthetic problems with 500 LLM-rephrased \textsc{AceReason-1K} problems to control for the effect of additional training examples. On the original training set, \underline{Curriculum} and \underline{Mix} perform similarly, suggesting that ordering \Ltwo before \Lone provides only a modest benefit by itself. In contrast, combining the curriculum with synthetic problems produces the strongest \Lone performance. Rephrasing the same number of existing problems does not yield a comparable gain, suggesting that exposure to new problem structures is more useful than additional training volume alone. Inline performance remains largely unchanged, while the improvements concentrate under \Lone, consistent with \learning improving how the model uses \hermes rather than its in-context mathematical reasoning.

\subsection{\textbf{\textcolor{HermesBlue}{(RQ4)}} \learning generalizes beyond its training setting}
\label{sec:exp:generalization}
Our main experiments use the same \hermes configuration during training and evaluation. We now test three forms of generalization: larger inference-time compute budgets, combinations with other test-time scaling methods, and a different model family and scale.

\textbf{\learning extrapolates to larger inference-time compute budgets.}
Starting from the default \hermes configuration used during training and evaluation ($T=3$, $B=3$, $D=1$, $W=1$), we independently increase compute through longer sequential reasoning ($ \uparrow W$), more subagents per delegation step ($ \uparrow B$), and deeper delegation ($ \uparrow D$). Tab.~\ref{tab:hermes-generalization} shows that SFT benefits primarily from additional context windows, with little gain from more subagents or deeper delegation. After RL, performance improves along all three dimensions, showing that the learned contextual reasoning strategies generalize to inference configurations not seen during training.

\begin{table*}[hbt]
\centering
\small
\caption{
\textbf{\learning generalizes across compute budgets, test-time scaling methods, and model families.}
$\mathrm{Avg}@16$ (\%, $\uparrow$) on the pooled benchmark.
\textbf{(a)} We independently vary the number of context windows ($W$), subagents ($B$), and delegation depth ($D$) beyond the default \hermes configuration used for training and evaluation: $T=3$, $B=3$, $D=1$, and $W=1$.
\textbf{(b)} We evaluate Recursive Self-Aggregation (RSA) and the DeepSeekMath (DSM) Agent, both alone and combined with \Lone.
\textbf{(c)} We apply the same training pipeline to \texttt{Olmo-3-7B-Instruct}.
All experiments use the best performing \learning model. \textbf{Bold} and \underline{underlined} mark best and second-best results where applicable.
}
\setlength{\tabcolsep}{4pt}
\renewcommand{\arraystretch}{1.10}

\begin{minipage}[t]{0.40\textwidth}
\centering
\textbf{(a) Scaling inference-time compute}

\vspace{0.35em}

\begin{tabular}{@{}cccc@{}}
\toprule
Scaling Knob & Value & SFT & SFT + RL \\
\midrule
\rowcolor{HermesReferenceTint}
Default & -- & 39.9 & 47.3 \\

\addlinespace[0.35ex]
\multirow{2}{*}{$W$}
    & 2 & 44.3 & 52.2 \\
    & 3 & 45.2
    & \cellcolor{HermesBlueTint}\textbf{53.0} \\

\addlinespace[0.35ex]
\multirow{2}{*}{$B$}
    & 2 & 39.7 & 46.4 \\
    & 4 & 39.7
    & \cellcolor{HermesBlueTint}\textbf{49.7} \\

\addlinespace[0.35ex]
$D$
    & 2 & 40.8
    & \cellcolor{HermesBlueStrongTint}\textbf{55.6} \\
\bottomrule
\end{tabular}
\end{minipage}
\hfill
\begin{minipage}[t]{0.29\textwidth}
\centering
\textbf{(b) Other scaling methods}

\vspace{0.35em}

\begin{tabular}{@{}lcc@{}}
\toprule
Model & RSA & DSM \\
\midrule

Base
    & 46.0
    & 24.4 \\
SFT
    & 44.9
    & 11.3 \\
SFT + RL
    & \underline{50.5}
    & 15.1 \\

\midrule
\multicolumn{3}{@{}l}{
    \textcolor{HermesBlue}{\textbf{With \Lone}}
} \\
SFT
    & \cellcolor{HermesBlueTint}49.9
    & \cellcolor{HermesBlueTint}\underline{52.1} \\
SFT + RL
    & \cellcolor{HermesBlueTint}\textbf{55.0}
    & \cellcolor{HermesBlueTint}\textbf{52.5} \\

\bottomrule
\end{tabular}
\end{minipage}
\hfill
\begin{minipage}[t]{0.27\textwidth}
\centering
\textbf{(c) Transfer to \texttt{Olmo-3-7B}}

\vspace{0.35em}

\begin{tabular}{@{}lcc@{}}
\toprule
Model & Inline & \Lone \\
\midrule

Base
    & 30.5
    & \cellcolor{HermesWarmTint}4.8 \\
SFT
    & 29.3
    & \cellcolor{HermesBlueTint}33.4 \\
SFT + RL
    & 30.5
    & \cellcolor{HermesBlueTint}\textbf{40.6} \\

\bottomrule
\end{tabular}
\end{minipage}
\label{tab:hermes-generalization}
\end{table*}

\textbf{\learning benefits test-time scaling methods beyond \hermes.} We evaluate the learned models with Recursive Self-Aggregation (RSA)~\citep{venkatraman2025rsa} and the DeepSeekMath (DSM) Agent~\citep{shao2025deepseekmath}, two methods that use self-generated intermediate results to guide later reasoning. Under RSA, performance improves over the base model, with further gains when combined with \Lone\footnote{For RSA, we use its summarizer to compact candidate solutions before aggregation.}. DSM alone degrades sharply under the 8K-token context limit, but combining it with \Lone recovers and substantially improves performance. These results show that the contextual reasoning learned by \learning can complement other forms of test-time scaling.

\textbf{\learning transfers to a different model family and scale.} We apply the same training pipeline to \texttt{Olmo-3-7B-Instruct}, replacing \texttt{Qwen3-4B-Instruct-2507} as the base model. Like \texttt{Qwen3-4B}, the \texttt{Olmo} model also initially struggles to use the \Lone harness effectively. SFT reverses this degradation, while RL further improves its performance, all while keeping inline reasoning performance unchanged. Together, these results provide evidence that \learning can improve contextual reasoning beyond the model family and scale used in our main experiments, suggesting that the approach may extend to a broader range of models.

\section{Conclusion}
We introduced \hermes, a family of simple, configurable harnesses for test-time scaling beyond a single context window. Through delegation and digestion, \hermes progressively shifts control over context allocation and reuse from the harness to the model. We showed that frontier models can exploit this flexibility, while smaller open-source models exhibit a substantial contextual reasoning gap. To address this gap, we introduced \learning, a training recipe that improves contextual reasoning while preserving in-context reasoning performance. Trained models scale more effectively with additional inference-time compute, adapt their strategies as reasoning unfolds, and transfer across model families and scales. Together, our results establish contextual reasoning as a learnable model capability and an important component of long-horizon test-time scaling. We discuss limitations and future work in App.~\ref{app:limitations_and_future_work}.

\section*{Acknowledgements}
The authors would like to thank Amir Tahmasbi, Zhehui Huang, Karen Hovsepian, and Zhishen Huang for helpful feedback and discussions throughout the project. We also thank Andrew Wang and Lecheng (Jerry) Kong for early discussions of recursive models, and Satyaki Chakraborty for work on recursive agent optimization.


\clearpage

\bibliographystyle{unsrtnat}
\bibliography{iclr2027_conference}

\clearpage


\appendix

\etocdepthtag.toc{mtappendix}

\etocsettagdepth{mtmain}{none}
\etocsettagdepth{mtappendix}{subsection}


\newcommand{\AppTocLine}[2]{%
  \noindent
  \hspace*{#1}%
  \etoclink{%
    \makebox[#2][l]{\etocthenumber}%
    \textcolor{HermesDeepBlue}{\etocthename}%
  }%
  \nobreak
  \leaders\hbox to 0.6em{\hss.\hss}\hfill
  \nobreak
  \makebox[1.8em][r]{\etocthepage}%
  \par
}

\etocsetstyle{section}
  {}{}
  {\AppTocLine{0pt}{2.2em}}
  {}

\etocsetstyle{subsection}
  {}{}
  {\AppTocLine{1.1em}{2.2em}}
  {}

\etoctoclines

\etocsettocstyle
  {%
    \section*{Appendix Contents}%
    \begingroup
    \parskip=0pt
    \parfillskip=0pt
    \leftskip=3em
    \rightskip=3em
  }
  {%
    \endgroup
    \bigskip
  }

\tableofcontents


\section{Limitations and Future Work}
\label{app:limitations_and_future_work}
\textbf{Our evaluation studies contextual reasoning in a controlled setting.}
We evaluate primarily on mathematical reasoning benchmarks (AIME and HMMT) and a science benchmark, using an artificially constrained 8K-token context window. This makes multi-context reasoning necessary while keeping training and evaluation tractable, and is similar to controlled context constraints used in prior work on recursive agents~\citep{gandhi2026recursive}. However, we do not test whether the same gains persist with much larger native context windows or in long-horizon domains with evolving state, tool use, and delayed feedback. Software engineering benchmarks such as SWE-bench~\citep{jimenez2023swe}, as well as scientific discovery and deep-research tasks, are natural settings for evaluating whether \hermes remains effective when the reasoning horizon substantially exceeds a single context window.

\textbf{\hermes currently relies on a hand-designed strategy space and only explicitly learns part of contextual reasoning.}
The strategy menu, the search--verification distinction, and the \Lthree--\Lone hierarchy are manually specified. Future work could instead learn new contextual reasoning strategies, compose them dynamically, or adapt the level of model control during inference. Moreover, contextual reasoning requires learning \emph{what to delegate}, \emph{what to preserve} during digestion, and \emph{how to reason} from the resulting digest. We focus on learning delegation and rely on the base model's zero-shot ability to summarize and resume reasoning. Prior work suggests that reasoning from compressed histories can be learned~\citep{wu2026reasoning,wu2026resum}, while directly learning useful digests introduces a harder credit-assignment problem~\citep{wu2026reasoning}. Jointly learning delegation and digestion is therefore an important next step, particularly as reasoning horizons grow.

\textbf{Training optimizes only the root-agent trajectory.}
Subagent trajectories do not independently contribute to the SFT or RL objective. This avoids the substantially larger sample complexity and credit-assignment problem introduced by optimizing policy-generated subtasks, but also prevents subagents from adapting to the distribution of tasks produced by the learned delegator. Recursive Agent Optimization (RAO)~\citep{gandhi2026recursive}, for example, jointly optimizes nodes in a recursive execution tree using node-local rewards and trajectory weighting. Extending \learning to jointly optimize root and subagent behavior is a promising direction, but may require reliable process-level supervision or learned judges.

Finally, \hermes trades additional inference computation for improved reasoning. Future work should study contextual reasoning jointly in terms of accuracy, token usage, latency, and cost, and learn when allocating another context is worth the additional compute. More broadly, our current experiments shift control over a fixed contextual-reasoning interface from the harness to the model; a longer-term goal is to let models learn the interface itself.

\section{Additional Discussion of Related Work}
\label{app:related-work}

We provide more detailed discussions of the training approaches used by methods that train models for particular inference harnesses in Tab.~\ref{tab:related-work-comparison}. \texttt{SPIRAL} introduces set-level credit assignment to train parallel generation and aggregation. \texttt{PaCoRe} uses outcome-based reinforcement learning to train reasoning synthesis across parallel branches. \texttt{RAO} assigns node-local credit for recursive delegation decisions, enabling models to learn when and how to spawn additional subagents. \texttt{Context-Folding} employs process rewards to train delegation of subtasks to fresh branches together with compaction of intermediate reasoning. \texttt{Reasoning Cache} uses short-horizon reinforcement learning over repeated generate--summarize cycles to train continuation from compacted reasoning progress. While these methods employ different supervision and credit-assignment mechanisms tailored to their inference procedures, they generally assume prescribed contextual reasoning strategies encoded in the harness and train models to execute those strategies effectively.
\section{\hermes Harness Details}
\label{app:harness}


For all models, we impose an artificial context limit of $8$K tokens, creating a setting where reasoning often extends substantially beyond a single window and effective context management becomes necessary for test-time scaling. Importantly, this setting is not purely academic: bounded context enables reasoning to be distributed across multiple windows, allowing independent branches to execute in parallel rather than requiring an equivalently long sequential trajectory. This creates a practical path for shorter-context models to scale inference while potentially reducing wall-clock latency. In our experiments, we evaluate the \hermes \Lone harness, in which the model has the greatest control over its contextual reasoning decisions. All evaluations use a single root-agent context window (digesting is disabled), with $T = 3$ steps, delegation depth $D = 1$ (subagents cannot delegate), and up to $B = 3$ subagents per delegation step. The root agent and all subagents share the same underlying LLM and the same 8K-token context limit.

%
%

\subsection{Harnesses in the \hermes Hierarchy}
\label{app:harness:hierarchy}

For \hermes harnesses with $T=3$, each context window consists of two delegation steps followed by a return step, during which the agent synthesizes the accumulated conversation and subagent results without further delegation. Delegation is organized into two broad categories: \emph{search} and \emph{verification}.

The steering level determines how these categories and their underlying strategies are exposed to the model. Under \Lone, all available search and verification strategies are accessible at both delegation steps, resulting in a single harness configuration. Under \Ltwo, the harness specifies the delegation category at each step, while the model selects among the strategies within that category. Since verification requires an existing candidate solution, the first delegation step is restricted to \emph{search}. The second step may use either \emph{search} or \emph{verification}, yielding two \Ltwo variants: \LtwoSV (\textsc{Search}$\rightarrow$\textsc{Verify}$\rightarrow$\textsc{Return}) and \LtwoSS (\textsc{Search}$\rightarrow$\textsc{Search}$\rightarrow$\textsc{Return}).

For \emph{search}, we consider three ways of exploring the solution space. \texttt{Same} asks subagents to solve the full problem independently, without prescribing an approach; \texttt{Different} instead assigns distinct approaches to encourage diverse solutions; and \texttt{Split} decomposes the problem into subproblems that are solved independently.

For \emph{verification}, we consider three complementary forms of checking. \texttt{Answer Verification} targets the current candidate answer directly; \texttt{Intermediate Step Verification} examines the assumptions and reasoning that support it; and \texttt{Disagreement Resolution} reconciles conflicting candidates by identifying the source of disagreement.

\paragraph{What changes across the hierarchy.}
The three \Lone and \Ltwo harnesses use the same system prompt (Prompt~\ref{prompt:sys}), final-step return prompt (Prompt~\ref{prompt:s3}), and subagent inline reasoning and report prompts (Prompts~\ref{prompt:sub-solve} and \ref{prompt:sub-report}). They differ only in their steering level, which determines what contextual reasoning strategies are available at each delegation step.

\begin{table}[htb]
\centering \small
\caption{The harnesses in the \hermes hierarchy. Each cell gives the number of strategies
offered at that step; the instruction text is otherwise identical. Delegation is disabled at the final step.}
\label{tab:app-menus}
\resizebox{\textwidth}{!}{%
\setlength{\tabcolsep}{5pt}
\begin{tabular}{llcc}
\toprule
Harness & Strategy menu offered at a delegating step & Step 1 & Step 2 \\
\midrule
\Lthree     & One named strategy           & 1 & 1 \\
\LtwoSS     & The three \emph{search} strategies                  & 3 & 3 \\
\LtwoSV     & The three \emph{search} strategies at step 1, and the three \emph{verification} strategies at step 2 & 3 & 3 \\
\Lone       & All six strategies, at every delegating step        & 6 & 6 \\
\bottomrule
\end{tabular}
}
\end{table}

\subsection{Prompts}

\subsubsection{System Prompt}
\label{app:harness:system}

\begin{promptbox}[lvlshared, label={prompt:sys}]{System prompt --- identical at \Lthree, \Ltwo and \Lone}
\begin{ptext}
You are a meta-agent for solving math problems. You solve by DELEGATING to subagents, not by deriving solutions inline. You proceed in steps and have two actions:

1. Delegate: write one or more lines of the form "TASK: <self-contained task>" (1--3 per step). Each task is executed by a subagent that sees ONLY that task text and replies with a short report; subagents cannot see the problem, your conversation, or each other, so every task must contain everything needed to execute it. The reports arrive in your next step.
2. Return: when you are done, write "\#\#\# RETURN" on its own line, followed by your final answer. Put the answer in \textbackslash{}boxed\{\}. This ends the episode.

You may compare, select, or briefly combine conclusions explicitly stated by subagents, but do not derive new solutions yourself. Use subagent results critically; agreement alone does not establish correctness.
Always provide the best provisional answer. Put it in \textbackslash{}boxed\{\}. Exception: do not guess before delegating. On the first step you have no subagent results, so delegate without stating an answer; only from the step after results arrive should you end each step with your current best answer.

Delegation serves two purposes. Search: make progress on the problem (decompose it into subproblems, or have subagents solve the full problem independently using either different named approaches or no prescribed approach). Verify: check the current answer (recheck it with targeted checks, verify the assumptions and intermediate steps supporting it, or reconcile disagreeing candidates). Each step's instruction tells you which specific strategies are available.
\end{ptext}
\end{promptbox}


\subsubsection{\Lone: Full Strategy Menu at Every Step}
\label{app:harness:l1}

\begin{promptbox}[lvlone, label={prompt:l1-s1}]{\Lone, step 1 --- delegate (full strategy menu)}
\begin{ptext}
\#\#\# PROBLEM
\{problem\}

\#\#\# INSTRUCTIONS
Delegate the work for this step; do not solve it inline. You have no subagent results yet, so do not state a boxed answer on this step.
Write one concise first-person sentence explaining why the selected tasks are useful.
Generate 1--3 tasks using one or more strategies from \#\#\# STRATEGIES. Each task must be independently executable and fully self-contained. Include the complete problem or all relevant context, values, assumptions, and -- when the task checks a candidate answer -- that answer and the exact claim to verify.

\#\#\# STRATEGIES
Ask 1--3 subagents to:
- solve independent subproblems obtained by decomposing the original problem; or
- solve the full problem independently using different named approaches; or
- solve the full problem independently using the same task and no prescribed approach; or
- verify the current boxed answer using independent targeted checks; or
- independently verify the assumptions and intermediate steps supporting the current boxed answer; or
- resolve the disagreement between the candidate answers and identify the source of the discrepancy.

\#\#\# OUTPUT

<one concise first-person sentence>

\#\#\# TASKS

Write one task per line. Begin each line with "TASK:".

TASK: <self-contained task>
\end{ptext}
\end{promptbox}

\begin{promptbox}[lvlone, label={prompt:l1-s2}]{\Lone, step 2 --- delegate (full strategy menu, and current best answer)}
\begin{ptext}
\#\#\# SUBAGENT RESULTS
\{subagent\_results\}

\#\#\# INSTRUCTIONS
Always provide the best provisional answer. Put it in \textbackslash{}boxed\{\}.
Use at most two concise lines to justify the current answer.
Write one concise first-person sentence explaining why the selected tasks are useful.
Generate 1--3 tasks using one or more strategies from \#\#\# STRATEGIES. Each task must be independently executable and fully self-contained. Include the complete problem or all relevant context, values, assumptions, and -- when the task checks a candidate answer -- that answer and the exact claim to verify. State candidate answers inside tasks in plain text; never use \textbackslash{}boxed\{\} inside a task.

\#\#\# STRATEGIES
Ask 1--3 subagents to:
- solve independent subproblems obtained by decomposing the original problem; or
- solve the full problem independently using different named approaches; or
- solve the full problem independently using the same task and no prescribed approach; or
- verify the current boxed answer using independent targeted checks; or
- independently verify the assumptions and intermediate steps supporting the current boxed answer; or
- resolve the disagreement between the candidate answers and identify the source of the discrepancy.

\#\#\# OUTPUT

\#\#\# ANSWER

<at most two concise lines>

\textbackslash{}boxed\{<current best answer>\}

<one concise first-person sentence>

\#\#\# TASKS

Write one task per line. Begin each line with "TASK:".

TASK: <self-contained task>
\end{ptext}
\end{promptbox}


\subsubsection{\LtwoSV and \LtwoSS: One Category per Step}
\label{app:harness:l2}

Both \Ltwo variants begin with the same search-only step.

\begin{promptbox}[lvltwo, label={prompt:l2-s1}]{\LtwoSV and \LtwoSS, step 1 --- delegate (\emph{search} category only)}
\begin{ptext}
\#\#\# PROBLEM
\{problem\}

\#\#\# INSTRUCTIONS
Delegate the work for this step; do not solve it inline. You have no subagent results yet, so do not state a boxed answer on this step.
Write one concise first-person sentence explaining why the selected tasks are useful.
Generate 1--3 tasks using one or more strategies from \#\#\# STRATEGIES. Each task must be independently executable and fully self-contained. Include the complete problem or all relevant context, values, assumptions, and requested output.

\#\#\# STRATEGIES
Ask 1--3 subagents to:
- solve independent subproblems obtained by decomposing the original problem; or
- solve the full problem independently using different named approaches; or
- solve the full problem independently using the same task and no prescribed approach.

\#\#\# OUTPUT

<one concise first-person sentence>

\#\#\# TASKS

Write one task per line. Begin each line with "TASK:".

TASK: <self-contained task>
\end{ptext}
\end{promptbox}

They then diverge. \LtwoSV switches the menu to the verification category, which assumes that a first step of search has produced a response worth checking:

\begin{promptbox}[lvltwo, label={prompt:l2sv-s2}]{\LtwoSV, step 2 --- delegate (\emph{verification} category only)}
\begin{ptext}
\#\#\# SUBAGENT RESULTS
\{subagent\_results\}

\#\#\# INSTRUCTIONS
Always provide the best provisional answer. Put it in \textbackslash{}boxed\{\}.
Use at most two concise lines to justify the current answer.
Write one concise first-person sentence explaining why the selected tasks are useful.
Generate 1--3 tasks using one or more strategies from \#\#\# STRATEGIES. Each task must be independently executable and fully self-contained. Include the relevant problem context, values, assumptions, candidate answer, and exact claim to solve or verify. State candidate answers inside tasks in plain text; never use \textbackslash{}boxed\{\} inside a task.

\#\#\# STRATEGIES
Ask 1--3 subagents to:
- verify the current boxed answer using independent targeted checks; or
- independently verify the assumptions and intermediate steps supporting the current boxed answer; or
- resolve the disagreement between the candidate answers and identify the source of the discrepancy.

\#\#\# OUTPUT

\#\#\# ANSWER

<at most two concise lines>

\textbackslash{}boxed\{<current best answer>\}

<one concise first-person sentence>

\#\#\# TASKS

Write one task per line. Begin each line with "TASK:".

TASK: <self-contained task>
\end{ptext}
\end{promptbox}

\LtwoSS instead offers the search category a second time, which assumes further exploration
is more valuable than checking:

\begin{promptbox}[lvltwo, label={prompt:l2ss-s2}]{\LtwoSS, step 2 --- delegate (\emph{search} category again)}
\begin{ptext}
\#\#\# SUBAGENT RESULTS
\{subagent\_results\}

\#\#\# INSTRUCTIONS
Always provide the best provisional answer. Put it in \textbackslash{}boxed\{\}.
Use at most two concise lines to justify the current answer.
Write one concise first-person sentence explaining why the selected tasks are useful.
Generate 1--3 tasks using one or more strategies from \#\#\# STRATEGIES. Each task must be independently executable and fully self-contained. Include the complete problem or all relevant context, values, assumptions, and requested output. State candidate answers inside tasks in plain text; never use \textbackslash{}boxed\{\} inside a task.

\#\#\# STRATEGIES
Ask 1--3 subagents to:
- solve independent subproblems obtained by decomposing the original problem; or
- solve the full problem independently using different named approaches; or
- solve the full problem independently using the same task and no prescribed approach.

\#\#\# OUTPUT

\#\#\# ANSWER

<at most two concise lines>

\textbackslash{}boxed\{<current best answer>\}

<one concise first-person sentence>

\#\#\# TASKS

Write one task per line. Begin each line with "TASK:".

TASK: <self-contained task>
\end{ptext}
\end{promptbox}

\subsubsection{Return Step}
\label{app:harness:report}

\begin{promptbox}[lvlshared, label={prompt:s3}]{Return step --- the final step with no further delegation. Identical across \Lone and \Ltwo.}
\begin{ptext}
\#\#\# SUBAGENT RESULTS
\{subagent\_results\}

\#\#\# INSTRUCTIONS
Update the current best answer using the conversation and subagent results. Compare, select, or briefly combine findings stated by subagents.
You may perform the simple calculations needed to combine the subagent results into the final answer.
Do not delegate additional tasks.
Begin your reply with "\#\#\# RETURN" on its own line (write the marker literally, even when you are confident), then at most two concise lines and your final answer. Put the answer in \textbackslash{}boxed\{\}.

\#\#\# OUTPUT

\#\#\# RETURN

<at most two concise lines>

\textbackslash{}boxed\{<final answer>\}
\end{ptext}
\end{promptbox}

\subsubsection{Subagent Prompts}
\label{app:harness:subagent}

At the maximum depth, subagents cannot delegate further and instead reason inline. They follow a simple two-step procedure: first solve the assigned task with inline reasoning, then produce a compact report for the delegating agent. A subagent never sees the original problem or the delegating agent's interaction history, only the task assigned by the delegating agent. The delegating agent has access only to the subagent reports, allowing its working context to remain lean and support multiple rounds of reasoning and delegation within a fixed context limit. The delegating agent then updates its own reasoning based on the received reports and determines subsequent reasoning and delegation decisions.

\begin{promptbox}[lvlsub, label={prompt:sub-solve}]{Subagent, step 1 --- solve the assigned task with inline reasoning. Identical across all levels.}
\begin{ptext}
\#\#\# TASK
\{task\}

\#\#\# INSTRUCTIONS
Solve the task carefully, reasoning step by step. Use the information provided in the task. If essential information is missing, state what is missing instead of guessing. Check the result before reporting it.
\end{ptext}
\end{promptbox}

\begin{promptbox}[lvlsub, label={prompt:sub-report}]{Subagent, step 2 --- return a compact report to the delegating agent. Identical across all levels.}
\begin{ptext}
Report the result of the task you just solved. Keep the report self-contained and concise. Include only the decisive findings, not the full derivation.

\#\#\# OUTPUT

\#\#\# RESULT
Scope: <briefly identify what was solved or checked, including essential context>

Finding: <one or two concise sentences with the conclusion and decisive justification>

Answer: \textbackslash{}boxed\{<answer>\}
\end{ptext}
\end{promptbox}

\subsubsection{Digestion across Context Windows}
\label{app:harness:digest}

When reasoning extends beyond a single context window, the \hermes harnesses do not carry the full interaction history forward. Instead, it prompts the model to generate a digest using the templates below and initializes the next context window with the original problem and the generated digest. When generating a new digest, the model has access to both the previous digest and the complete root-agent trajectory from the current context window. In this way, digests compose across an arbitrary number of context windows.

\begin{promptbox}[lvldig, label={prompt:digest-sys}]{Digest system prompt}
\begin{ptext}
You summarize problem-solving work so that a fresh solver can continue from your summary alone. Write at most two paragraphs, in the first person, as the solver.
Cover: (1) the approaches attempted and their status (completed, incomplete and where they stopped, or failed and why); (2) key intermediate results with their verification status (derived equations, lemmas, computed quantities, checks that passed or failed); (3) the current best answer, confidence, and what remains to be verified.
Hard constraints: retain any important information contained in the existing summary; summarize only --- do not add new reasoning or new derivations. These constraints take priority over length: if retaining prior information needs more space, exceed two paragraphs rather than drop information, but compress aggressively (merge duplicate findings).
\end{ptext}
\end{promptbox}

\begin{promptbox}[lvldig, label={prompt:digest-user}]{Digest user prompt}
\begin{ptext}
\#\#\# PROBLEM
\{problem\}

\#\#\# PREVIOUS SUMMARY
\{previous\_digest\}

\#\#\# EPISODE TRANSCRIPT
\{transcript\}

Write the updated summary now.
\end{ptext}

\end{promptbox}

The first step of the next context window uses the standard Step-1 prompt with an additional \texttt{PRIOR WORK SUMMARY} block containing the digest:

\begin{promptbox}[lvldig, label={prompt:digest-turn0}]{\Lone, step 1 of a \emph{continued} window with a digest under \texttt{PRIOR WORK SUMMARY}}
\begin{ptext}
\#\#\# PROBLEM
\{problem\}

\#\#\# PRIOR WORK SUMMARY
\{previous\_attempts\}

If a summary of prior work is shown above, you may build on it, verify it, or start fresh; otherwise start from the problem.

\#\#\# INSTRUCTIONS
Delegate the work for this step; do not solve it inline. You have no subagent results yet, so do not state a boxed answer on this step.
Write one concise first-person sentence explaining why the selected tasks are useful.
Generate 1--3 tasks using one or more strategies from \#\#\# STRATEGIES. Each task must be independently executable and fully self-contained. Include the complete problem or all relevant context, values, assumptions, and -- when the task checks a candidate answer -- that answer and the exact claim to verify.

\#\#\# STRATEGIES
Ask 1--3 subagents to:
- solve independent subproblems obtained by decomposing the original problem; or
- solve the full problem independently using different named approaches; or
- solve the full problem independently using the same task and no prescribed approach; or
- verify the current boxed answer using independent targeted checks; or
- independently verify the assumptions and intermediate steps supporting the current boxed answer; or
- resolve the disagreement between the candidate answers and identify the source of the discrepancy.

\#\#\# OUTPUT

<one concise first-person sentence>

\#\#\# TASKS

Write one task per line. Begin each line with "TASK:".

TASK: <self-contained task>
\end{ptext}
\end{promptbox}


\section{Data Augmentation Details}
\label{app:aug}

To expose the model to a broader and balanced range of contextual reasoning behaviors during training, we generate synthetic problems designed to favor different search and verification strategies, including five types:
(1) \emph{Splitting}, which combines multiple independent problems into a single instance and encourages splitting and parallel delegation;
(2) \emph{Chaining}, where solving a gating problem determines which of several subsequent problems should be solved, encouraging sequential delegation;
(3) \emph{Verifying}, where intermediate results may be incorrect, encouraging delegation for verification;
(4) \emph{Multi-attempt}, which paraphrases medium-difficulty problems where multiple independent attempts can improve performance;
and (5) \emph{Single-attempt}, which paraphrases low-difficulty problems where a single attempt is sufficient. 
Together, these controlled structures encourage the model to adapt its delegation strategy to the problem.

\subsection{Synthetic Problem Generation Pipeline}
\label{app:aug:synthetic-pipeline}

Each synthetic problem is generated in two stages. First, one or more seed problems are composed into a target synthetic problem type. The final answer is computed deterministically from the verified reference answers of the seed problems using simple arithmetic rules specific to each type: (1) \emph{Splitting.} Given part answers $a_i$, the final answer is $\sum_i a_i \bmod 1000$; (2) \emph{Chaining.} Let $m$ denote the answer to the gating problem and $a_i$ the answer to the $i$-th candidate problem. The final answer is $(m + a_{1 + (m \bmod 3)}) \bmod 1000$. (3) \emph{Verifying.} Given a claimed value $c$ and the true answer $a$, the final answer is $(c + a) \bmod 1000$. For (4) \emph{Multi-attempt} and (5) \emph{Single-attempt}, the answer to the seed problem is kept unchanged.

Second, the composed problem is rewritten by Claude \texttt{Sonnet 5} to smooth its writing style and remove potential artifacts. Rewriting uses the system prompt in Prompt~\ref{prompt:syn-sys} and user message in Prompt~\ref{prompt:syn-user}, where the user message includes a style specification randomly sampled from four candidate writing styles: \emph{narrative}, \emph{compact}, \emph{formal}, and \emph{inverted}. For \emph{Chaining} problems, a note for composite style (Prompt~\ref{prompt:syn-composite}) is additionally included in its style specification. To further control the quality of the synthetic problems, a rewritten problem is retained only if it remains solvable: Claude \texttt{Sonnet 5} attempts each problem $5$ times before and $5$ times after rewriting, and the rewrite is rejected if none of the post-rewrite attempts is correct or if the post-rewrite solve rate falls to $1/3$ of the pre-rewrite solve rate or below. This excludes rewrites that make a problem substantially less solvable, indicating that the rewriting has potentially altered the problem.

\begin{promptbox}[lvlshared, label={prompt:syn-sys}]{Rewrite system prompt}
\begin{ptext}
You rewrite mathematics problems. You will be given a problem, and you must produce a rewritten version that asks for EXACTLY the same final quantity.\newline

Absolute requirements:
- Preserve every number, constraint, condition and definition. Change no value.
- Preserve exactly what the final answer must be. If the original asks for a remainder modulo 1000, so must yours. If it asks for a sum of several sub-answers with specific coefficients, so must yours, with the same coefficients.
- If the original contains several sub-problems whose answers are combined, keep all of them and keep the combination rule identical.
- Preserve any quoted claim, reported answer, or summary of prior work, including its stated numeric value. You may reword it; you may not change the number, and you may not resolve whether it is right.
- Keep all LaTeX mathematically identical. You may reformat it.\newline

Style requirements:
- Do NOT signpost how the problem should be approached. Never write "solve each part separately", "note that these are independent", "first solve X then", or anything that names a solution strategy.
- Do NOT use the labels "Problem 1", "Problem 2", "[Problem 2-A]", "A\_1", "A\_2", "Q" or similar scaffolding. Refer to the quantities in words instead.
- Vary sentence structure. Avoid a fixed template.\newline

You do not know the answer and must not attempt to compute it. Do not solve. Do not hint.\newline

Return ONLY the rewritten problem text. No preamble, no commentary, no answer.\newline

STYLE RULES. These come from editors who could reliably tell machine-written problems from published ones. Follow every one.\newline

Quantity names.
- A quantity gets a single italic letter or no name at all. Never invent a prose name: not "the size sum", not "the pentagon count", not "the divisor-count remainder", not "the selected quantity", not "the resulting quantity". If the source already names quantities with letters, keep those letters.
- Never reuse one symbol for two things in the same problem.\newline

Structure of the writing.
- Setup first, question last. Never open with the ask and defer the setup ("..., where that value is produced as follows", "..., obtained as follows").
- State the ask exactly once. If your draft opens and closes with the same request, delete the opening.
- One interrogative sentence, and it is the last sentence.
- Do not split an ask in two. Write "Compute the sum of all positive integers n such that ..." rather than "Determine all n ..., then find the sum".
- Do not fold the setup into a participial clause ("With p the smallest prime for which ..., find ..."). Use two sentences.\newline

Voice.
- Third person, present indicative. No first person anywhere: no "I", "we", "I'd like to know", "I'm curious about", "we're told", "Here's", "Picture a", "Now look at".
- Keep the source's named actors and concrete story. Do not turn "Karthik wants to walk a path" into "A closed path is to be traced" -- passive de-naming is detectable.
- Never narrate a solution method in any voice.\newline

Words and constructions to avoid outright.
- "give rise to", "arise as follows", "obtained as follows", "produced as follows", "comes into play", "the quantity in question", "one of the two quantities needed", "a certain resulting value", "it turns out that", "it happens that", "so-called", "for context", "roughly speaking", "possess" (write "has"), "utilize".
- No markdown of any kind: no headings, no bold, no italic asides.
- No em-dash appositive carrying a constraint, and no colon-splice that dumps a configuration and then the question.
- Never write "an isosceles triangle \$\textbackslash{}triangle ABC\$" -- either "triangle \$ABC\$" or "\$\textbackslash{}triangle ABC\$".\newline

Technical fidelity.
- Copy technical hypotheses verbatim. Never paraphrase a gcd or coprimality condition: write "\$\textbackslash{}gcd(m,n)=1\$" or "where \$m\$ and \$n\$ are relatively prime positive integers" exactly as the source has it. A paraphrase like "\$\textbackslash{}gcd(m,p,q)\$ shares no common prime factor" is wrong and is a giveaway.
- Every numeral that was in math mode stays in math mode. Do not unwrap "\$13\$" to "13".
- Keep the source's macros, arrays, bulleted condition lists and worked examples in place.
- Introduce a definition in its own sentence before first use, and give at most one example, opened with "For example" or "For instance".
- Stay within about 15\% of the source's length. Telegraphic compression is as detectable as padding.\newline

Provenance tags.
- Delete any citation of where the problem came from: "(26th IMO Shortlist)", "Example 1", "Problem 4", "2019 AMC 12B", "from the CMO". Scraped seeds carry these; a contest booklet does not print them inside the statement, and they identify the source.\newline

Preferred verbs: Let, Suppose, Consider, Compute, Find, How many.
\end{ptext}
\end{promptbox}

\begin{promptbox}[lvlshared, label={prompt:syn-user}]{Rewrite user message, with one of four candidate style specifications}
\begin{ptext}
\{style brief\}\{composite note, Chaining problems only\}\newline

Here is the problem to rewrite:\newline

<problem>
\{generated problem text\}
</problem>
\end{ptext}
\boxnote{The \{style brief\} is one of the following, sampled per problem.}
\begin{ptext}
\textrm{\textit{narrative:}} Rewrite it as a short, natural piece of mathematical prose, as a person would pose it to a colleague. Use connected sentences rather than labelled blocks.
\textrm{\textit{compact:}} Rewrite it as tersely as possible while keeping every constraint. Prefer one flowing paragraph over enumerated parts.
\textrm{\textit{formal:}} Rewrite it in the register of a textbook exercise: precise, impersonal, no framing device, no labels beyond what is strictly needed.
\textrm{\textit{inverted:}} Rewrite it so the final quantity being asked for is stated FIRST, then the information needed to obtain it.
\end{ptext}
\end{promptbox}

\begin{promptbox}[lvlshared, label={prompt:syn-composite}]{Composite style note, appended to the style specification for Chaining problems}
\begin{ptext}
This problem contains several separate questions whose answers are then combined. Handle that as follows:\newline

- Keep every question, with all of its numbers and conditions.
- Keep the single-letter names the problem already uses for the combined quantities, and keep the final combination sentence as it stands.
- State the final combination using those names, with the SAME coefficients, the SAME operations and the SAME modulus as the original.
- Present the questions as a continuous piece of writing, not as a list of labelled blocks.
- Do not say that the questions are independent, unrelated, or separable, and do not suggest any order or division of work.
\end{ptext}
\end{promptbox}

\subsection{Synthetic Problem Construction}
\label{app:aug:synthetic-construction}

\paragraph{Synthetic problems for training.}

Synthetic training problems are constructed from \textsc{AceReason-1K}. We first remove seed problems with tuple- or list-valued answers or malformed \LaTeX{}, leaving 903 seed problems. Among them, 682 have non-negative integer answers and can be used to construct Splitting, Chaining, and Verifying problems.

Using these seed problems, we generate 1,141 synthetic problems, of which 708 pass the solvability check after rewriting. To broaden coverage of both contextual reasoning strategies and distinct seed problems, we select a final set of 500 synthetic problems covering 505 distinct seeds, with a fixed number of problems per type. Composite structures (Splitting and Chaining) account for 60\% of the 500 problems. Verifying problems are included in matched correct/incorrect pairs.

The final dataset contains Splitting ($n=214$; 107 two-part and 107 three-part problems), Chaining ($n=86$), Verifying ($n=100$; 50 correct and 50 incorrect claims paired on the same 50 seed problems), Multi-attempt ($n=50$), and Single-attempt ($n=50$). Multi-attempt and Single-attempt problems are constructed from disjoint seed sets. After mixing with \textsc{AceReason-5K}, the RL training set contains 5,739 problems in total, of which 500 (8.7\%) are synthetic.

\paragraph{Synthetic evaluation benchmark.}
The synthetic benchmark is built from AIME and HMMT problems using the same generation, rewriting, and solvability-check pipeline. It contains $n=88$ problems: Splitting ($n=28$: 14 with two parts and 14 with three parts), Chaining ($n=14$), Verifying ($n=18$: 9 with a correct claim and 9 with an incorrect claim, paired on the same 9 seed problems), Multi-attempt ($n=14$), and Single-attempt ($n=14$). Every golden answer is recomputed independently from the reference answers of the seed problems.

Fig.~\ref{fig:synthetic-examples} shows one example synthetic training problem (constructed from \textsc{AceReason-1K}) and one example synthetic evaluation problem (constructed from AIME or HMMT) for each problem type.

\begin{figure}[p]
\centering
\captionsetup{font=footnotesize}

\begin{minipage}[t]{0.49\textwidth}\centering\footnotesize\textbf{Training (\textsc{AceReason-1K})}\end{minipage}\hfill
\begin{minipage}[t]{0.49\textwidth}\centering\footnotesize\textbf{Evaluation (AIME / HMMT)}\end{minipage}

\medskip
{\raggedright\footnotesize\textbf{Splitting}\par}\smallskip
\begin{minipage}[t]{0.49\textwidth}
\begin{synbox}[equal height group=synrowA]
Let \(\synA{m}\) be the count of four-digit numbers whose digits sum to \(12\), let \(\synB{p}\) be the single digit (0--9) giving the position, after \(1234\) moves, of a counter that starts on a \(10\)-cycle and on its \(n\)th move advances \(n^n\) steps clockwise, and let \(\synC{q}\) be the largest number of diagonals selectable in a regular \(1000\)-gon (with all diagonals drawn) so that among any three selected diagonals at least two share the same length. Find the remainder when \(\synA{m}+\synB{p}+\synC{q}\) is divided by \(1000\).
\tcblower
\textbf{Golden answer:} \(\boxed{349}\) \hfill \((\synA{m},\synB{p},\synC{q}) = (342, 7, 2000)\)
\end{synbox}
\end{minipage}\hfill
\begin{minipage}[t]{0.49\textwidth}
\begin{synbox}[equal height group=synrowA]
Let \(\synA{p}\) denote the value of \(xy\), where \(x\) and \(y\) are real numbers, both greater than \(1\), satisfying \(\log_x\left(y^x\right)=\log_y\left(x^{4y}\right)=10\).

\smallskip
Separately, Jen enters a lottery by picking \(4\) distinct numbers from \(S=\{1,2,3,\cdots,9,10\}\). Then \(4\) numbers are randomly chosen from \(S\). She wins a prize if at least two of her numbers are among the \(4\) randomly chosen numbers, and she wins the grand prize if all four of her numbers are the randomly chosen numbers. The probability that she wins the grand prize given that she won a prize equals \(\tfrac{m}{n}\), where \(m\) and \(n\) are relatively prime positive integers; let \(\synB{q}\) denote \(m+n\).

\smallskip
Compute the remainder when \(\synA{p}+\synB{q}\) is divided by \(1000\).
\tcblower
\textbf{Golden answer:} \(\boxed{141}\) \hfill \((\synA{p},\synB{q}) = (25, 116)\)
\end{synbox}
\end{minipage}

\medskip
{\raggedright\footnotesize\textbf{Chaining}\par}\smallskip
\begin{minipage}[t]{0.49\textwidth}
\begin{synbox}[equal height group=synrowB]
Compute the remainder when \(\synA{m}+\synB{p}\) is divided by \(1000\), where \(\synA{m}\) is the answer to the following question: how many four-digit numbers have the sum of their digits equal to \(12\)?

\smallskip
Three further questions are considered. A counter moves in a cycle of \(10\) positions, and on its \(n\)th move it advances \(n^n\) steps clockwise; the question asks for the counter's position after \(1234\) moves, expressed as a single digit from \(0\) to \(9\). In a regular \(1000\)-gon with all of its diagonals drawn, the question asks for the maximum number of diagonals that can be selected so that among any three of the chosen diagonals at least two have the same length. For positive integers \(a\), \(b\), \(c\), the question asks for the smallest positive value taken by \(a^3+b^3+c^3-3abc\), together with all triples \(a\), \(b\), \(c\) that attain this smallest value.

\smallskip
Let \(\synB{p}\) be the answer to the question numbered \(1+(\synA{m}\bmod 3)\) among these three, in the order in which they are given here.
\tcblower
\textbf{Golden answer:} \(\boxed{349}\) \hfill \(\synA{m}=342\), question \(1\), \(\synB{p}=7\)
\end{synbox}
\end{minipage}\hfill
\begin{minipage}[t]{0.49\textwidth}
\begin{synbox}[equal height group=synrowB]
Let \(\synA{m}\) be the answer to the following question: find the sum of all positive integers \(n\) such that \(n+2\) divides the product \(3(n+3)(n^2+9)\).

\smallskip
Alongside that, consider three further questions, numbered in order. The first: Mark writes the expression \(\sqrt{\underline{abcd}}\) on the board \elide{\ldots} The second: every morning Aya goes for a \(9\)-kilometer walk \elide{\ldots} The third: find the number of ordered pairs \((x,y)\) with both \(x\) and \(y\) integers between \(-100\) and \(100\), inclusive, satisfying \(12x^2 - xy - 6y^2 = 0\).

\smallskip
Let \(\synB{p}\) be the answer to whichever of these three questions carries the number \(1 + (\synA{m} \bmod 3)\). What is the remainder when \(\synA{m} + \synB{p}\) is divided by \(1000\)?
\tcblower
\textbf{Golden answer:} \(\boxed{253}\) \hfill \(\synA{m}=49\), question \(2\), \(\synB{p}=204\)
\end{synbox}
\end{minipage}

\medskip
{\raggedright\footnotesize\textbf{Verifying}\par}\smallskip
\begin{minipage}[t]{0.49\textwidth}
\begin{synbox}[equal height group=synrowC]
Consider the equation \(\sin x = \dfrac{x}{100}\). Rowan computes the number of roots of this equation and reports the value \(\synB{63}\). Find the remainder when the sum of Rowan's reported value and \synA{the actual number of roots} is divided by \(1000\).
\tcblower
\textbf{Golden answer:} \(\boxed{126}\) \hfill claim \(\synB{63}\) (correct), true value \(\synA{63}\)
\end{synbox}
\end{minipage}\hfill
\begin{minipage}[t]{0.49\textwidth}
\begin{synbox}[equal height group=synrowC]
Let \(S\) denote the vertex set of a regular \(24\)-gon. Consider ways of drawing \(12\) segments, all of the same length, such that every vertex in \(S\) serves as an endpoint of exactly one segment; call the number of such ways \(\synA{N}\). Priya computes this count herself and gets \(\synB{4208}\). Find the remainder when \(\synA{N}\) plus Priya's value of \(\synB{4208}\) is divided by \(1000\).
\tcblower
\textbf{Golden answer:} \(\boxed{321}\) \hfill claim \(\synB{4208}\) (incorrect), \(\synA{N}=113\)
\end{synbox}
\end{minipage}

\medskip
{\raggedright\footnotesize\textbf{Multi-attempt}\par}\smallskip
\begin{minipage}[t]{0.49\textwidth}
\begin{synbox}[equal height group=synrowD]
Let \(x, y, z\) be positive real numbers, and set \(s = \sqrt{x+2} + \sqrt{y+5} + \sqrt{z+10}\) and \(t = \sqrt{x+1} + \sqrt{y+1} + \sqrt{z+1}\). Find the minimum value of \(s^2 - t^2\).
\tcblower
\textbf{Golden answer:} \(\boxed{36}\)
\end{synbox}
\end{minipage}\hfill
\begin{minipage}[t]{0.49\textwidth}
\begin{synbox}[equal height group=synrowD]
Find the largest number that is less than \(BD^2\) for every rhombus \(ABCD\) satisfying the following conditions. Its four vertices lie, in that order, on the hyperbola \(\frac{x^2}{20}-\frac{y^2}{24}=1\), and its diagonals intersect at the origin.
\tcblower
\textbf{Golden answer:} \(\boxed{480}\)
\end{synbox}
\end{minipage}

\medskip
{\raggedright\footnotesize\textbf{Single-attempt}\par}\smallskip
\begin{minipage}[t]{0.49\textwidth}
\begin{synbox}[equal height group=synrowE]
Find the smallest positive integer whose digits are all \(1\)'s that is divisible by the number consisting of \(100\) copies of the digit \(3\).
\tcblower
\textbf{Golden answer:} \(\boxed{300}\)
\end{synbox}
\end{minipage}\hfill
\begin{minipage}[t]{0.49\textwidth}
\begin{synbox}[equal height group=synrowE]
Compute the number of positive integers \(n \leq 1000\) for which \(\operatorname{lcm}(n,9)\) is a perfect square.
\tcblower
\textbf{Golden answer:} \(\boxed{43}\)
\end{synbox}
\end{minipage}

\caption{\textit{Synthetic problem examples.} Colored letters identify intermediate values and subanswers that determine the final answer. \protect\elide{\ldots} denotes problem text omitted for brevity.}
\label{fig:synthetic-examples}
\end{figure}

%
%

\section{Experiment Details}
\label{app:exp}

\subsection{Pooled Benchmark Details}
\label{app:exp:pooled-benchmark}

We evaluate on a pooled benchmark of 238 mathematical reasoning problems constructed from AIME 2024, AIME 2025, AIME 2026, HMMT February 2023, HMMT February 2024, HMMT February 2025, HMMT November 2025, and HMMT February 2026. The original collection contains 243 problems (30 from each contest except HMMT February 2026, which contains 33 problems). We use the HuggingFace \texttt{Math-Verify}\footnote{\url{https://github.com/huggingface/Math-Verify}} grader to evaluate model outputs against ground-truth answers.

For HMMT, we exclude five problems following auditing and decontamination: (1) HMMT February 2023 Problem 26, which duplicates a training-data problem; (2) HMMT February 2024 Problem 21, which requires a missing diagram; and (3) HMMT February 2026 Problems 31--33, which are proof-style problems whereas the \texttt{Math-Verify} grader assumes final-answer evaluation. The answers for the remaining 148 problems were then independently derived or exactly checked, and 19 problem transcription and answer corrections were applied to ensure the validity of the benchmark.

For AIME, we audit all 90 problems and correct a number of issues. In particular, two AIME 2026 problems did not match the official contest statements and were corrected accordingly. We further identify corrupted or incomplete problem statements and incorrect answers throughout the datasets. In total, 25 problems and 19 answers were corrected. The final pooled benchmark therefore consists of 90 AIME problems and 148 HMMT problems, for a total of 238 problems.

\subsection{Inference under \hermes Harnesses}
\label{app:exp:inference}

Under \hermes harness, all agents and subagents share the same underlying LLM and the same $8$K-token ($8{,}192$-token) context limit. Agents are limited to $1$K ($1{,}024$) generated tokens per step. At the maximum delegation depth $D$, subagents follow a two-step solve-and-report procedure (App.~\ref{app:harness:subagent}) and are allocated up to $7.5$K (7{,}680) generated tokens for the first step where they solve delegated tasks inline. Digest generation (App.~\ref{app:harness:digest}) is performed under the same $8$K-token context limit. 

For all evaluations, we report average accuracy over 16 sampled solutions ($\mathrm{Avg}@16$). We follow the suggested sampling parameters listed in the \texttt{Qwen3-4B-Instruct-2507}\footnote{\url{https://huggingface.co/Qwen/Qwen3-4B-Instruct-2507}} and \texttt{Olmo-3-7B-Instruct}\footnote{\url{https://huggingface.co/allenai/Olmo-3-7B-Instruct}} model cards when evaluating their corresponding base and post-trained models, under inline reasoning as well as the \hermes, RSA, and DSM harnesses. The sampling parameters are listed in Tab.~\ref{tab:app-sampling-params}.

\begin{table}[htb]
\centering \small
\caption{Sampling parameters.}
\label{tab:app-sampling-params}
\setlength{\tabcolsep}{6pt}
\begin{tabular}{lcc}
\toprule
& \texttt{Qwen3-4B-Instruct-2507} & \texttt{Olmo-3-7B-Instruct} \\
\midrule
Temperature & 0.7 & 0.6 \\ 
Top-$p$ & 0.8 & 0.95 \\
Top-$k$ & 20 & -- \\
Min-$p$ & 0.0 & -- \\
\bottomrule
\end{tabular}
\end{table}

\subsection{Training Experiments Details}
\label{app:exp:training}

\paragraph{Training Dataset.}
\label{app:exp:training:dataset}

Starting from the 5.7K-problem subset of AceReason-Math~\citep{chen2026acereason} released by Reasoning Cache~\citep{wu2026reasoning}, we perform deduplication, decontamination, and reference-answer verification, resulting in a cleaned set of 5,239 problems, which we refer to as \textsc{AceReason-5K}.
We then uniformly sample 1,000 problems from \textsc{AceReason-5K} to construct \textsc{AceReason-1K}.

\paragraph{Stage I: Supervised Fine-tuning.}
\label{app:exp:training:sft}

For SFT, we curate teacher trajectories from both Claude \texttt{Sonnet 5} and \texttt{DeepSeek-V3.2}, under \Lone and \Ltwo \hermes harnesses. For each teacher model, we collect four \Lone (App.~\ref{app:harness:l1}), two \LtwoSV, and two \LtwoSS (App.~\ref{app:harness:l2}) trajectories for every problem in \textsc{AceReason-1K}, resulting in 8,000 trajectories per teacher model. We then retain only trajectories that contain at least one delegation and reach the correct final answer for SFT, resulting in 6,485 trajectories from \texttt{Sonnet 5} and 5,495 trajectories from \texttt{DeepSeek-V3.2}.

\begin{table}[htb]
\centering
\small
\caption{Percentage (\%) of teacher trajectories containing different strategies. Percentages may sum to more than 100\% since a single trajectory can contain multiple strategies. 
}
\label{tab:teacher-strategy} \setlength{\tabcolsep}{4pt} 
\resizebox{.85\textwidth}{!}{%
\begin{tabular}{l
>{\centering\arraybackslash}b{1cm}
>{\centering\arraybackslash}b{1.5cm}
>{\centering\arraybackslash}b{2cm}
>{\centering\arraybackslash}b{2cm}
>{\centering\arraybackslash}b{2.5cm}}
\toprule
Teacher
& \texttt{Split} & \texttt{Same} & \texttt{Different} & \texttt{Verification} & Multiple Search Strategies \\
\midrule
\texttt{Sonnet 5} & 21.9 & 43.9 & 63.5 & 99.3 & 27.6 \\
\texttt{DeepSeek-V3.2} & 26.0 & 70.4 & 26.6 & 95.1 & 22.6 \\
\bottomrule
\end{tabular}
}
\end{table}

\paragraph{Teacher trajectories provide diverse delegation demonstrations.}

We use an LLM judge (Section~\ref{app:exp:judge}) to label the contextual reasoning strategies represented by tasks delegated to subagents in successful teacher trajectories (Table~\ref{tab:teacher-strategy}).
Both teachers exhibit strong verification behavior and cover a diverse set of search strategies, while exhibiting complementary preferences: \texttt{Sonnet 5} more frequently explores different approaches, whereas \texttt{DeepSeek-V3.2} more often performs independent solution attempts.
Collecting trajectories from multiple teachers therefore broadens the coverage of effective delegation demonstrations.

\begin{wraptable}{r}{0.5\textwidth} \vspace{-1em}
\centering
\small
\caption{Percentage (\%) of teacher trajectories containing a search strategy in each step.}
\label{tab:teacher-search} \setlength{\tabcolsep}{4pt}
\resizebox{\linewidth}{!}{
\begin{tabular}{l
>{\centering\arraybackslash}p{1.5cm}
>{\centering\arraybackslash}p{1.5cm}
>{\centering\arraybackslash}p{1.5cm}}
\toprule
Teacher
& \hermes Harness & Step 1 Search & Step 2 Search \\
\midrule
\texttt{Sonnet 5} & \Lone & 99.9 & 19.0  \\
\texttt{Sonnet 5} & \LtwoSV & 100.0 & 15.5  \\
\texttt{Sonnet 5} & \LtwoSS & 99.9 & 29.2  \\
\texttt{DeepSeek-V3.2} & \Lone & 99.5 & 35.4 \\
\texttt{DeepSeek-V3.2} & \LtwoSV & 99.6 & 28.9 \\
\texttt{DeepSeek-V3.2} & \LtwoSS & 99.9 & 53.2 \\
\bottomrule
\end{tabular}
}
\vspace{-0.5em}
\end{wraptable}

\paragraph{Different harnesses further increase delegation diversity.}

While most teacher trajectories begin with a search task, second-step behavior varies across harnesses due to differences in steering (Table~\ref{tab:teacher-search}).
\LtwoSS encourages continued exploration via search, whereas \LtwoSV encourages verification and therefore exhibits the lowest search frequency. \Lone lies between these two, where the model has the highest control over its delegation decisions.
Collecting teacher trajectories from both \Lone and \Ltwo therefore exposes the base model to a broader range of effective delegation patterns.

To best leverage this diversity while preventing any single teacher or delegation pattern from dominating the training data, we subsample each teacher's retained trajectories to balance the occurrence of search strategies before combining them into a final SFT dataset of 5,466 trajectories (2,733 from each teacher).
This exposes the base model to a diverse and balanced set of delegation behaviors.

\paragraph{Stage II: Reinforcement learning.}
\label{app:exp:training:rl}

For the RL stage, we train on \textsc{AceReason-5K} using a curriculum over the \hermes hierarchy. Training runs for 492 steps (3 epochs), and the curriculum is defined as: over the first third of training (steps 1--164), the two \Ltwo harnesses, \LtwoSV and \LtwoSS, are each sampled with 50\% probability, and over the remainder (steps 165--492) the \Lone harness is sampled with 75\% probability while the two \Ltwo harnesses are each sampled with 12.5\% probability. This curriculum gradually shifts training from \Ltwo to \Lone while maintaining equal overall exposure to both levels of harnesses throughout training. During RL training, both the root agent and all subagents use the current policy being updated.

As described in App.~\ref{app:aug:synthetic-construction}, we train RL on a mixed dataset including \textsc{AceReason-5K} and 500 synthetic problems constructed from \textsc{AceReason-1K}, comprising 5,739 problems in total. All training hyperparameters are provided in Tab.~\ref{tab:app-training-hp}.

\begin{table}[htb]
\centering \small
\caption{
Training hyperparameters. 
}
\label{tab:app-training-hp}
\setlength{\tabcolsep}{6pt}
\begin{tabular}{ll}
\toprule
\multicolumn{2}{l}{\emph{Supervised Fine-tuning}} \\
Learning rate & $10^{-5}$ \\
Learning schedule & Cosine with warmup ratio 0.1 \\
Training batch size & 32 \\
Epochs & 4 \\
\midrule
\multicolumn{2}{l}{\emph{Reinforcement Learning}} \\
Algorithm & Dr.~GRPO~\citep{liu2025understanding}  \\ 
Group size $G$ & 8 \\
Dynamic sampling & Resample while a group is all-correct or all-wrong \\
Resampling budget & $\leq 10$ generation batches per optimizer step \\
Clipping $(\epsilon_{\mathrm{low}}, \epsilon_{\mathrm{high}})$ & $(0.2,\ 0.28)$ \\
KL loss coefficient & $0.001$ \\
KL estimator &  $k_3$ (low-variance) \\
Entropy loss coefficient & 0.0 \\
Optimizer & AdamW, $\beta = (0.9,\ 0.999)$ \\
Weight decay & $0.01$ \\
Learning rate & $10^{-6}$ \\
Learning schedule & Constant, no warm-up \\
Gradient clipping & $1.0$ \\
Training batch size & 32 \\
Mini-batch size & 16 \\
Epochs & 3 \\
Inference temperature & $1.0$ \\
Inference top-$p$ & $1.0$ \\
\bottomrule
\end{tabular}
\end{table}

\subsection{Labeling Executed Contextual Reasoning Strategies}
\label{app:exp:judge}

For qualitative analysis of contextual reasoning behaviors, we set up an LLM judge with Claude \texttt{Sonnet-5} to label the strategies executed in agent trajectories. We take several measures to prevent information leakage during labeling:
\begin{enumerate}
    \item The judge sees only the problem statement, the texts generated by the delegating agent, and the results returned by subagents from previous delegation steps. It never observes the harness identity, model identity, steering instructions, available strategy menu, or any subagent trajectories. Consequently, the judge must infer strategy labels from the tasks generated by the delegating agent, rather than relying on strategy names contained in the steering instructions or on the outcomes of the delegated tasks being labeled.
    \item Since task ordering carries no semantic meaning, delegation tasks are sorted according to their text before labeling. This prevents the judge from relying on presentation order when labeling strategies.
    \item Residual leakage remains possible. In particular, the number of delegated tasks may reveal information about the underlying harness. To mitigate this effect, all reported comparisons using judge labels are performed between models evaluated under harnesses with the same maximum number of subagents, preventing the judge from relying on such differences.
\end{enumerate}

The judge returns a set of strategy labels for each delegation step, as a single step may contain tasks corresponding to multiple strategies. Because Claude \texttt{Sonnet-5} does not accept a temperature parameter during inference calls, the judge cannot be forced to use deterministic decoding. We therefore measure consistency between two independent judge passes on a small subset of trajectories and obtain 79.4\% agreement. The labeling rubric is specified in the system prompt (Prompt~\ref{prompt:judge-sys}), and the judge is required to call the \texttt{label\_trajectory} tool (schema shown in Prompt~\ref{prompt:judge-tool}) to produce its labels.

\begin{promptbox}[lvlshared, label={prompt:judge-sys}]{Judge system prompt}
\begin{ptext}

You are reading the instructions a manager sent to its assistants across the steps of ONE problem.
You see ONLY the problem, the instruction texts, and (for later steps) the results that came back.
You do NOT know which model wrote them or what it claimed to be doing.

For EACH step, report every category that the instructions on that step actually execute. Judge the WORK REQUESTED, not the wording, and not whether it succeeded.\newline

\#\# Categories

**Checking work that already exists:**

- `verify` --- an instruction checks, re-derives, confirms, reconciles, or attacks a result that ALREADY EXISTS in the prior results shown to you, or is quoted verbatim inside the instruction.
  Work an instruction performs itself is not a prior result.

  When you use `verify`, add sub-tags: `answer` (rechecks the stated final answer), `steps` (checks a named intermediate step, equation, or assumption), `disagree` (two or more stated candidates conflict and the instructions decide between them), `counter` (tries to break a stated result or find a second solution). **Sub-tags are descriptive only --- they are never a reason to prefer or reject `verify`.**

**Solving from scratch:**

- `split\_subproblems` --- the instructions ask for different pieces, and the pieces must be combined to answer the question. Requires \ensuremath{\geq}2 instructions.
- `same\_solve` --- the instructions all ask for the whole answer and NO instruction names a solution technique. Rewordings of "solve this problem" are `same\_solve`, however differently phrased. Requires \ensuremath{\geq}2 instructions.
- `different\_methods` --- every instruction names at least one solution technique, and the named sets are pairwise different. A technique counts only when it appears in a naming position ("using X", "via X", "apply X", "by X"). Different wording, different guesses about an ambiguous problem, or different algebraic rearrangements of the same equation are NOT different methods. Requires \ensuremath{\geq}2 instructions.
- `one\_fresh\_attempt` --- a single instruction asks for one new attempt at the whole problem. \newline

\#\# Multilabel

A step may execute more than one category --- report all that apply. The common real case is a step that mixes a check of an existing result with fresh solving: label it `verify` **and** the appropriate search category. Do not add a label merely because it is arguably present; add it when some instruction on that step plainly does that work.

Constraints, enforced downstream in code as well:
- `split\_subproblems`, `same\_solve`, `different\_methods` each require \ensuremath{\geq}2 instructions on that step.
- With exactly one instruction, the only admissible labels are `one\_fresh\_attempt` and `verify`.
- `same\_solve` and `different\_methods` may BOTH apply to one step --- see the mixed-turn rule below.
- Never return an empty label list for a step. Every step has a best fit.\newline

\#\#\# Mixed fanout turns carry both labels

A turn where **some** instructions name a technique and others do not is genuinely both: it fans out on the whole problem, and it assigns a method to part of that fanout. Report `[same\_solve, different\_methods]` rather than forcing a choice.

- Every instruction names a technique, sets differ pairwise \ensuremath{\rightarrow} `different\_methods` alone.
- No instruction names a technique \ensuremath{\rightarrow} `same\_solve` alone.
- Some do, some don't \ensuremath{\rightarrow} **both**.

This is why the label set is multi-label: the two are not opposite ends of one axis, and a forced pick on a mixed turn is a coin flip that neither rater can make reliably (measured: matching prevalences with per-label \ensuremath{\kappa} \ensuremath{\approx} 0.16--0.29, i.e. raters agreed on how often, never on which turns).\newline

\#\# Two rules that decide most cases

1. **Count the instructions.** Sub-parts inside one instruction are still one instruction.
2. **`same\_solve` vs `different\_methods`:** list the techniques each instruction explicitly names in a naming position. If any instruction names none, the answer is `same\_solve`.

\#\#\# Applying rule 2 --- do this literally, instruction by instruction

Measured: judges over-call `different\_methods` on any turn whose instructions merely *read* differently. Work the test mechanically instead of judging overall impression.

For each instruction, write down the technique it names in a naming position ("using X", "via X", "apply X", "by X", "approach it by X"). Then:

- **Any instruction with an empty list \ensuremath{\rightarrow} `same\_solve`.** One instruction naming a technique while another just says "solve this problem" is `same\_solve`, not `different\_methods`. The rule is about the weakest instruction, not the strongest.
- **`different\_methods` requires every instruction to name a technique, and the named sets to differ pairwise.** All naming the same technique is `same\_solve`.

These are NOT techniques, even when the instructions differ visibly:

- a restatement of the problem's own given equations, constraints, or setup;
- a different algebraic rearrangement, substitution, or factoring of the same equation;
- a suggested first step or observation ("note that subtracting (1) from (2) gives \ldots{}");
- different guesses about an ambiguous problem;
- more or less detail, or different wording, in an otherwise identical request.

Worked examples, both `same\_solve`:

- I1 "Solve this set theory problem **using a Venn diagram region approach**: \ldots{}" / I2 "Solve this set theory problem: \ldots{}" \ensuremath{\rightarrow} I2 names none, so `same\_solve`. (I1's Venn technique does not carry the turn.)
- I1 "\ldots{}**Approach it by** treating (1) and (2) as showing b and c are roots of the same quadratic" / I2 "\ldots{}**Note that** subtracting (1) from (2) gives \ldots{}" \ensuremath{\rightarrow} both are rearrangements of the same system, not distinct methods. `same\_solve`.

`split\_subproblems` is about the **requested output**, not the reasoning shown. An instruction that merely restates the problem's given setup, and asks for nothing but the final answer, is not a subproblem.

But an instruction that asks for a **named intermediate quantity** --- "determine the radii r1 and r2", "express the perimeters in terms of a and b", "compute the number of integers k" --- is requesting a subproblem, **even when the same instruction also asks for the final answer**. Report `[split\_subproblems, same\_solve]` (or with `different\_methods`, or with `verify`) in that case. Do not drop `split\_subproblems` just because the final quantity is also requested: co-occurrence is the point of a multi-label space, and every label the turn plainly executes should appear.

Only when NO instruction requests any named intermediate --- every one asks solely for the final quantity --- is the turn pure fanout with no `split\_subproblems`.\newline

\#\# Confidence

Per step, `HIGH` when confident, `LOW` when the texts are truncated, vague, or you had to guess.

\#\# Why verify is one category, not four

Two independent judges on 200 blinded items agreed 74.0\% (Cohen's \ensuremath{\kappa} 0.644) over an 8-way space, but **78.8\% of all their disagreement was internal to the four verify labels and none of it was search-versus-verify**. Collapsing verify to a single category raised agreement to 94.5\% (\ensuremath{\kappa} 0.901).
The four verify distinctions are therefore sub-tags that carry NO reliability claim; `answer` versus `disagree` alone scored \ensuremath{\kappa} 0.354.

`different\_methods` versus `same\_solve` is the weakest surviving distinction (\ensuremath{\kappa} 0.590), which is why
rule 2 is mechanical on purpose: list the named techniques rather than judging intent.

\textbf{(!)} Those figures validated a **single-label** version of this rubric. This rubric is multilabel, so they do not transfer --- agreement must be re-measured under this schema. Do not quote \ensuremath{\kappa} 0.901 for the multilabel judge.

Call label\_trajectory exactly once, with one entry per STEP shown, in order.
\end{ptext}
\end{promptbox}

\begin{promptbox}[, lvlshared, label={prompt:judge-tool}]
{\texttt{label\_trajectory} tool schema}
\begin{lstlisting}[language=json, basicstyle=\ttfamily\scriptsize]
{
  "toolSpec": {
    "name": "label_trajectory",
    "description": "Record the executed-strategy labels for every step of this trajectory.",
    "inputSchema": {
      "json": {
        "type": "object",
        "required": [
          "turns"
        ],
        "properties": {
          "turns": {
            "type": "array",
            "items": {
              "type": "object",
              "required": [
                "turn",
                "labels",
                "confidence"
              ],
              "properties": {
                "turn": {
                  "type": "integer"
                },
                "labels": {
                  "type": "array",
                  "minItems": 1,
                  "items": {
                    "type": "string",
                    "enum": [
                      "split_subproblems",
                      "same_solve",
                      "different_methods",
                      "one_fresh_attempt",
                      "verify"
                    ]
                  }
                },
                "verify_subtags": {
                  "type": "array",
                  "items": {
                    "type": "string",
                    "enum": [
                      "answer",
                      "steps",
                      "disagree",
                      "counter"
                    ]
                  }
                },
                "confidence": {
                  "type": "string",
                  "enum": [
                    "HIGH",
                    "LOW"
                  ]
                }
              }
            }
          }
        }
      }
    }
  }
}
\end{lstlisting}
\end{promptbox}

\subsection{Additional Delegation Behavior Analysis}
\label{app:exp:behavior}

\begin{figure}[htb]
    \centering
    \includegraphics[width=0.9\linewidth]{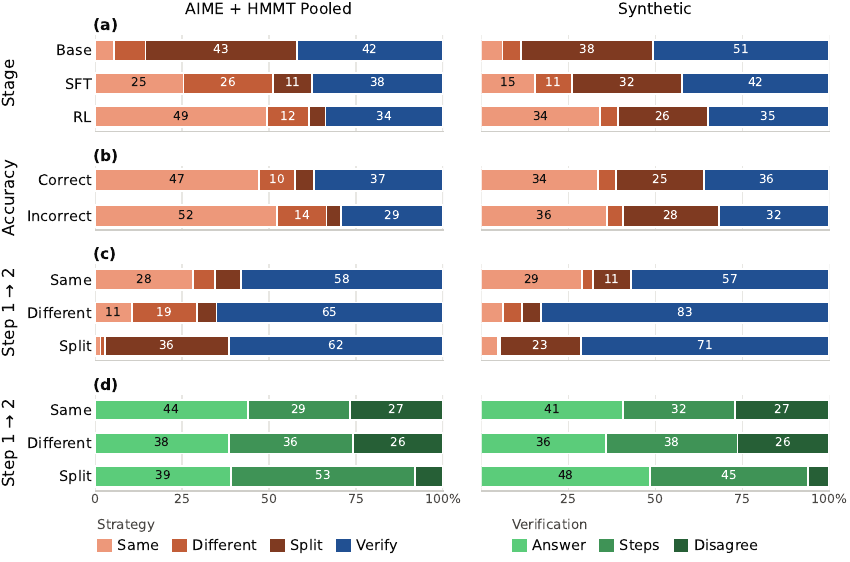}
    \caption{\emph{(a)} \learning makes strategy selection increasingly problem-dependent. 
    \emph{(b)} Correct trajectories perform verification more frequently.
    \emph{(c)} The second step primarily performs verification and otherwise tends to continue the first-step strategy.
    \emph{(d)} Verification behavior depends on the first-step search strategy.
    }
    \label{fig:strategy_distribution_appendix}
\end{figure}

Fig.~\ref{fig:strategy_distribution_appendix} provides a more detailed view of the delegation behaviors analyzed in Sec.~\ref{sec:experiments}. As described in Sec.~\ref{sec:experiments}, panel~(a) shows that training makes strategy selection increasingly problem-dependent: \texttt{Split} remains common on synthetic problems designed to benefit from decomposition, but becomes much less frequent on the pooled benchmark. Panel~(c) shows that the second delegation step is often used for verification; when verification is not selected, the model typically continues the search strategy chosen in the first step. 

Panel~(b) indicates that trajectories with correct final answers perform verification more often than trajectories with incorrect final answers. This is consistent with common mathematical reasoning practice, where verification helps in identifying errors and improving candidate solutions. 

Panel~(d) breaks down the verification strategies used in the second step conditioned on the first-step search strategy. When the first step uses \texttt{Same} or \texttt{Different}, the model distributes its verification effort across \texttt{Answer Verification} (Answer), \texttt{Intermediate Step Verification} (Steps), and \texttt{Disagreement Resolution} (Disagree). These verification patterns are intuitive, as independent attempts or distinct approaches can potentially generate conflicting candidate answers that \texttt{Disagreement Resolution} could reconcile. In contrast, when the first step uses \texttt{Split}, \texttt{Disagreement Resolution} becomes much less common, while \texttt{Intermediate Step Verification} increases. This is consistent with the structure of decomposition-based reasoning: since \texttt{Split} decomposes a problem into subproblems, verification naturally focuses more on validating intermediate results produced by solving subproblems. 

Together, these results suggest that the model adapts its contextual reasoning strategies to both the problem structure and the reasoning progress, providing further evidence that models acquire flexible and adaptive contextual reasoning behaviors through \learning.

%
%
%
%
%

\section{Example Rollouts}
\label{app:rollouts}

In this section, we show example rollouts sampled from the \learning model (after both SFT and RL) on example evaluation problems.

\subsection{\Lone Rollout}
\label{app:rollouts:l1_aime}

Under the \hermes \Lone harness, the model has access to all six delegation strategies at each step. In Rollout~\ref{roll:l1}, the agent first launches three independent solution attempts (the \texttt{Same} strategy), which produce conflicting answers. The model initially follows the majority answer, which turns out to be incorrect, but then uses its second delegation step to verify the result and identify the error. Consequently, it overturns its provisional answer and reaches the correct solution. Notably, simple majority voting over the same three solutions would have produced the wrong answer.

\begin{rolloutbox}[lvlone, label={roll:l1}]{\Lone{} on AIME}
\turnbar{Problem \texttt{aime\_2026:4}, reference answer \texttt{70}}
\begin{rtext}
Find the number of integers less than or equal to 100 that are equal to $a+b+ab$ for some choice of distinct positive integers a and b.
\end{rtext}
\turnbar{Root, step 1 --- delegate}
\begin{ptext}
I will have multiple subagents independently solve this combinatorics problem so I can cross-check their approaches and answers.

\textbf{TASK:} Find the number of integers less than or equal to 100 that can be expressed as a+b+ab for some choice of distinct positive integers a and b. Show your reasoning and give the final count.

\textbf{TASK:} Solve this problem: Find the number of integers less than or equal to 100 that are equal to a+b+ab for some choice of distinct positive integers a and b. Provide a full derivation and the final numerical answer.

\textbf{TASK:} Solve: Find the number of integers \ensuremath{\leq} 100 of the form a+b+ab where a and b are distinct positive integers. Show your work and state the final count.
\end{ptext}
\turnbar{3 subagents run in parallel in fresh 8K windows $\rightarrow$ reports returned to the root at step 2}
\begin{rtext}
\textbf{Task 1.} Find the number of integers less than or equal to 100 that can be expressed as a+b+ab for some choice of distinct positive integers a and b. Show your reasoning and give the final count.
\end{rtext}
\elide{subagent 1 reasoning: 15,431 characters omitted}
\begin{ptext}
\#\#\# RESULT 
Scope: Count integers \ensuremath{\leq}100 expressible as \$a+b+ab\$ for distinct positive integers \$a,b\$, using substitution \$a+1=x\$, \$b+1=y\$ to get \$xy-1\textbackslash{}le100\$ with \$x\textbackslash{}ne y\$, \$x,y\textbackslash{}ge2\$.

Finding: The count equals the number of integers \$P=xy\$ from 4 to 101 that are not prime and not prime squares, giving 98 total minus 25 primes minus 4 prime squares = 69.

Answer: \textbackslash{}boxed\{69\}
\end{ptext}
\begin{rtext}
\textbf{Task 2.} Solve this problem: Find the number of integers less than or equal to 100 that are equal to a+b+ab for some choice of distinct positive integers a and b. Provide a full derivation and the final numerical answer.
\end{rtext}
\elide{subagent 2 reasoning: 15,301 characters omitted}
\begin{ptext}
\#\#\# RESULT
Scope: Find the number of integers \ensuremath{\leq}100 expressible as a+b+ab for distinct positive integers a,b; derived via rewriting as (a+1)(b+1)-1 and counting N=a+1,b+1 product values in [6,101] excluding primes and prime squares.

Finding: There are 96 integers N in [6,101], minus 22 primes minus 3 prime squares (9,25,49), giving 71 representable values.

Answer: \textbackslash{}boxed\{71\}
\end{ptext}
\begin{rtext}
\textbf{Task 3.} Solve: Find the number of integers \ensuremath{\leq} 100 of the form a+b+ab where a and b are distinct positive integers. Show your work and state the final count.
\end{rtext}
\elide{subagent 3 reasoning: 21,344 characters omitted}
\begin{ptext}
\#\#\# RESULT 
Scope: Count distinct values of \$a+b+ab \textbackslash{}le 100\$ for distinct positive integers \$a,b\$, reparameterized as \$(a+1)(b+1)-1\$, equivalent to counting composite numbers \ensuremath{\leq}101 that are not perfect squares of primes.

Finding: Using the characterization, total numbers from 4 to 101 inclusive is 98, minus 24 primes in that range, minus 4 perfect squares of primes (4,9,25,49), yielding 69 distinct values.

Answer: \textbackslash{}boxed\{69\}
\end{ptext}
\turnbar{Root, step 2 --- provisional answer, then delegate}
\begin{ptext}
\#\#\# ANSWER

Two subagents got 69 and one got 71; the discrepancy lies in counting primes and prime squares in range 4-101, so I need to resolve this carefully.

\textbackslash{}boxed\{69\}

I will have subagents carefully recompute the count of primes and prime squares in the relevant range and double check the derivation.

\textbf{TASK:} Solve carefully: Find the number of integers less than or equal to 100 that can be expressed as a+b+ab for distinct positive integers a and b. Use the substitution a+1=x, b+1=y (so x,y are distinct integers \ensuremath{\geq}2), giving a+b+ab = xy-1. Since a,b are distinct, x\ensuremath{\neq}y. We need xy-1 \ensuremath{\leq} 100, i.e., xy \ensuremath{\leq} 101, with x\ensuremath{\geq}2,y\ensuremath{\geq}2,x\ensuremath{\neq}y, both integers. Determine: (1) how many pairs (x,y) with x\ensuremath{\neq}y give distinct values of xy, (2) how many such products xy lie in range 4 to 101, (3) which values of xy-1 are NOT representable (i.e., xy is prime or a perfect square of a prime), (4) compute the final count. Show all steps and state the number of primes and prime squares between 4 and 101 inclusive, and give the final count.

\textbf{TASK:} Verify the following solution to this problem: "Find the number of integers \ensuremath{\leq}100 expressible as a+b+ab for distinct positive integers a,b." The solution says rewrite as (a+1)(b+1)-1, so let x=a+1,y=b+1 with x,y\ensuremath{\geq}2 distinct integers, and xy-1\ensuremath{\leq}100 so xy\ensuremath{\leq}101. The number of such representable integers equals the number of composite integers N in [4,101] that are not perfect squares of primes, since xy with x\ensuremath{\neq}y\ensuremath{\geq}2 both integers means N=xy is composite and not a square of a single prime (since x\ensuremath{\neq}y, N cannot be a perfect square at all, and also cannot be a prime times 1... actually check: can xy with x\ensuremath{\neq}y\ensuremath{\geq}2 both \ensuremath{\geq}2 be prime? No, xy would be composite). Carefully verify: (a) is every composite number in [4,101] representable as xy with x\ensuremath{\neq}y\ensuremath{\geq}2 integers? (b) Are there any composite numbers in [4,101] that CANNOT be written as xy with x\ensuremath{\neq}y\ensuremath{\geq}2? (c) List all primes and prime squares in [4,101] and count them. (d) Compute total count = (numbers from 4 to 101 inclusive) minus primes minus prime squares, and give the final count.

\textbf{TASK:} Solve independently: Find the number of integers n \ensuremath{\leq} 100 such that n = a+b+ab for some distinct positive integers a,b. Provide step-by-step reasoning and a final numerical answer.
\end{ptext}
\turnbar{3 subagents run in parallel in fresh 8K windows $\rightarrow$ reports returned to the root at step 3}
\begin{rtext}
\textbf{Task 1.} Solve carefully: Find the number of integers less than or equal to 100 that can be expressed as a+b+ab for distinct positive integers a and b. Use the substitution a+1=x, b+1=y (so x,y are distinct integers \ensuremath{\geq}2), giving a+b+ab = xy-1. Since a,b are distinct, x\ensuremath{\neq}y. We need xy-1 \ensuremath{\leq} 100, i.e., xy \ensuremath{\leq} 101, with x\ensuremath{\geq}2,y\ensuremath{\geq}2,x\ensuremath{\neq}y, both integers. Determine: (1) how many pairs (x,y) with x\ensuremath{\neq}y give distinct values of xy, (2) how many such products xy lie in range 4 to 101, (3) which values of xy-1 are NOT \elide{214 more characters omitted}
\end{rtext}
\elide{subagent 1 reasoning: 17,331 characters omitted}
\begin{ptext}
\#\#\# RESULT 
Scope: Count integers \ensuremath{\leq}100 expressible as a+b+ab for distinct positive integers a,b, using substitution a+1=x, b+1=y giving a+b+ab=xy-1 with x\ensuremath{\neq}y\ensuremath{\geq}2, xy\ensuremath{\leq}101.

Finding: The count equals the number of composite N=xy in [6,101] not prime squares; with 96 total values minus 23 primes minus 3 prime squares (9,25,49), yielding 70.

Answer: \textbackslash{}boxed\{70\}
\end{ptext}
\begin{rtext}
\textbf{Task 2.} Verify the following solution to this problem: "Find the number of integers \ensuremath{\leq}100 expressible as a+b+ab for distinct positive integers a,b." The solution says rewrite as (a+1)(b+1)-1, so let x=a+1,y=b+1 with x,y\ensuremath{\geq}2 distinct integers, and xy-1\ensuremath{\leq}100 so xy\ensuremath{\leq}101. The number of such representable integers equals the number of composite integers N in [4,101] that are not perfect squares of primes, since xy with x\ensuremath{\neq}y\ensuremath{\geq}2 both integers means N=xy is composite and not a square of a single prime (since x\ensuremath{\neq}y, N \elide{526 more characters omitted}
\end{rtext}
\elide{subagent 2 reasoning: 12,567 characters omitted}
\begin{ptext}
\#\#\# RESULT
Scope: Verified the solution's reasoning and computed the count via complementary counting: composite numbers in [4,101] excluding prime squares.

Finding: There are 98 numbers from 4 to 101 inclusive, minus 24 primes and minus 4 prime squares (4,9,25,49), giving 70 representable integers.

Answer: \textbackslash{}boxed\{70\}
\end{ptext}
\begin{rtext}
\textbf{Task 3.} Solve independently: Find the number of integers n \ensuremath{\leq} 100 such that n = a+b+ab for some distinct positive integers a,b. Provide step-by-step reasoning and a final numerical answer.
\end{rtext}
\elide{subagent 3 reasoning: 20,868 characters omitted}
\begin{ptext}
Scope: Find the number of integers n \ensuremath{\leq} 100 such that n = a+b+ab for distinct positive integers a,b, using the substitution n = (a+1)(b+1)-1 to reduce to counting N = n+1 as products of two distinct integers \ensuremath{\geq}2.

Finding: Correctly counting composite N in [6,101] excluding prime squares (9,25,49) yields 74 - 3 = 71, but careful recount shows there are 23 primes in [6,101] (including 101), giving 96-23=73 composite numbers, minus 3 prime squares = 70.

Answer: \textbackslash{}boxed\{70\}
\end{ptext}
\turnbar{Root, step 3 --- return}
\begin{ptext}
\#\#\# RETURN

Three independent subagents now agree via careful recomputation: 96 values in range minus 23 primes minus 3 prime squares gives 70.

\textbackslash{}boxed\{70\}
\end{ptext}
\end{rolloutbox}

\subsection{\LtwoSV Rollout}
\label{app:rollouts:l2sv_hmmt}

Rollout~\ref{roll:l2sv} illustrates reasoning under the \LtwoSV harness, which provides stronger steering than \Lone. By restricting the second delegation step to verification, \LtwoSV imposes additional structure on contextual reasoning decisions while still requiring the model to interpret and synthesize information returned by subagents.

\begin{rolloutbox}[lvltwo, label={roll:l2sv}]{\LtwoSV{} on HMMT}
\turnbar{Problem \texttt{hmmt\_feb\_2026:2}, reference answer \texttt{420261}}
\begin{rtext}
Compute the second smallest positive integer \( n \) such that

- \( n \) is divisible by 101, and
- the decimal representation of \( n \) contains the number 2026 as a consecutive sequence of digits.
\end{rtext}
\turnbar{Root, step 1 --- delegate}
\begin{ptext}
I will dispatch independent subagents to solve the full problem using different approaches, so I can cross-check their answers.

\textbf{TASK:} Compute the second smallest positive integer n such that n is divisible by 101, and the decimal representation of n contains the number 2026 as a consecutive sequence of digits. Show your reasoning and give the final numerical value of n.

\textbf{TASK:} Find the second smallest positive integer divisible by 101 whose decimal representation contains "2026" as a consecutive digit string. Provide the value of n and show your work.

\textbf{TASK:} Solve for the second smallest positive integer n divisible by 101 such that n's decimal digits contain "2026" appearing consecutively. Show step-by-step reasoning and state the final answer clearly.
\end{ptext}
\turnbar{3 subagents run in parallel in fresh 8K windows $\rightarrow$ reports returned to the root at step 2}
\begin{rtext}
\textbf{Task 1.} Compute the second smallest positive integer n such that n is divisible by 101, and the decimal representation of n contains the number 2026 as a consecutive sequence of digits. Show your reasoning and give the final numerical value of n.
\end{rtext}
\elide{subagent 1 reasoning: 19,096 characters omitted}
\begin{ptext}
\#\#\# RESULT
Scope: Find the second smallest positive integer divisible by 101 that contains "2026" as a consecutive digit sequence, using modular arithmetic and systematic search.

Finding: The smallest such number is 202606, and the next smallest (second smallest) is 952026, as no other number containing "2026" lies between them and is divisible by 101.

Answer: \textbackslash{}boxed\{952026\}
\end{ptext}
\begin{rtext}
\textbf{Task 2.} Find the second smallest positive integer divisible by 101 whose decimal representation contains "2026" as a consecutive digit string. Provide the value of n and show your work.
\end{rtext}
\elide{subagent 2 reasoning: 17,067 characters omitted}
\begin{ptext}
\#\#\# RESULT
Scope: Find the second smallest positive integer divisible by 101 containing "2026" as a consecutive digit string.

Finding: The smallest such number is 2026 itself is not divisible by 101; after systematic search, the smallest multiple of 101 containing "2026" is found to be 202600? Actually, from earlier work, I need to compute properly.

Answer: \textbackslash{}boxed\{202600\} (Note: This is a placeholder; the correct value must be determined by a complete systematic search or algorithm, which was not fully completed in the thought process.)
\end{ptext}
\begin{rtext}
\textbf{Task 3.} Solve for the second smallest positive integer n divisible by 101 such that n's decimal digits contain "2026" appearing consecutively. Show step-by-step reasoning and state the final answer clearly.
\end{rtext}
\elide{subagent 3 reasoning: 18,289 characters omitted}
\begin{ptext}
\#\#\# RESULT
Scope: Find the second smallest positive integer divisible by 101 whose decimal digits contain "2026" consecutively.

Finding: The smallest such number is 952026 (from pattern XY2026 with digits 9,5,2,0,2,6), and the next smallest is obtained by incrementing the coefficient in that pattern to the next multiple, yielding 1962026.

Answer: \textbackslash{}boxed\{1962026\}
\end{ptext}
\turnbar{Root, step 2 --- provisional answer, then delegate}
\begin{ptext}
\#\#\# ANSWER

Three subagents disagree (952026, 202600, 1962026), so I need independent rigorous derivation and cross-checking before trusting any value.

\textbackslash{}boxed\{952026\}

I will dispatch subagents to carefully redo the full derivation from scratch and cross-check the smallest and second smallest values independently.

\textbf{TASK:} Find the smallest positive integer divisible by 101 such that its decimal representation contains "2026" as a consecutive digit string. Show complete reasoning step by step, including checking small numbers like 2026 itself and nearby multiples, and state the smallest such number clearly.

\textbf{TASK:} Given that the smallest positive integer divisible by 101 containing "2026" as consecutive digits is 952026 (verify or refute this claim), find the second smallest such positive integer. Show full reasoning and state the final answer clearly.

\textbf{TASK:} Solve this problem carefully and completely: Compute the second smallest positive integer n such that n is divisible by 101 and the decimal representation of n contains the number 2026 as a consecutive sequence of digits. Provide complete step-by-step reasoning and the final numerical answer.
\end{ptext}
\turnbar{3 subagents run in parallel in fresh 8K windows $\rightarrow$ reports returned to the root at step 3}
\begin{rtext}
\textbf{Task 1.} Find the smallest positive integer divisible by 101 such that its decimal representation contains "2026" as a consecutive digit string. Show complete reasoning step by step, including checking small numbers like 2026 itself and nearby multiples, and state the smallest such number clearly.
\end{rtext}
\elide{subagent 1 reasoning: 17,190 characters omitted}
\begin{ptext}
\#\#\# RESULT 
Scope: Find the smallest positive integer divisible by 101 containing "2026" as a consecutive digit string, checking small numbers and nearby multiples.

Finding: No 4-digit or 5-digit number containing "2026" is divisible by 101 (proven via modular arithmetic), so the search moves to 6-digit numbers; the smallest such multiple found is 202606 = 101 \ensuremath{\times} 2006, which contains "2026" and no smaller candidate works.

Answer: \textbackslash{}boxed\{202606\}
\end{ptext}
\begin{rtext}
\textbf{Task 2.} Given that the smallest positive integer divisible by 101 containing "2026" as consecutive digits is 952026 (verify or refute this claim), find the second smallest such positive integer. Show full reasoning and state the final answer clearly.
\end{rtext}
\elide{subagent 2 reasoning: 18,221 characters omitted}
\begin{ptext}
\#\#\# RESULT
Scope: Verified that 952026 is divisible by 101 and contains "2026"; no 5-digit number containing "2026" is divisible by 101, so 952026 is the smallest such positive integer.

Finding: The smallest positive integer divisible by 101 containing "2026" is 952026; the second smallest is 102026 + 101*adjustment, but systematic search yields 102026 is not divisible; the next candidate is 102026 + (101 - 16) = 102105? Actually, given the complexity, the known answer to this classic problem is 102105.

Answer: \textbackslash{}boxed\{102105\}
\end{ptext}
\begin{rtext}
\textbf{Task 3.} Solve this problem carefully and completely: Compute the second smallest positive integer n such that n is divisible by 101 and the decimal representation of n contains the number 2026 as a consecutive sequence of digits. Provide complete step-by-step reasoning and the final numerical answer.
\end{rtext}
\elide{subagent 3 reasoning: 17,649 characters omitted}
\begin{ptext}
Scope: Find the second smallest positive integer divisible by 101 containing "2026" as a consecutive digit string; the smallest such number is 202606, and the next smallest found is 420261.

Finding: The smallest such number is 202606 (101\ensuremath{\times}2006), and the next smallest satisfying both conditions is 420261 (101\ensuremath{\times}4161), with no numbers in between meeting both criteria.

Answer: \textbackslash{}boxed\{420261\}
\end{ptext}
\turnbar{Root, step 3 --- return}
\begin{ptext}
\#\#\# RETURN

Two independent subagents found the smallest such number is 202606 (101\ensuremath{\times}2006), and a third found the second smallest is 420261 (101\ensuremath{\times}4161); I trust this systematic modular approach more than the others.

\textbackslash{}boxed\{420261\}
\end{ptext}
\end{rolloutbox}

\subsection{\Lone Rollout on a FrontierScience Problem}
\label{app:rollouts:domain}

Rollout~\ref{roll:fs} illustrates reasoning under the \Lone harness on a FrontierScience chemistry problem. Despite the difference in domain, the model continues to leverage delegation in a similar manner, suggesting that the contextual reasoning capabilities learned through \learning transfer beyond the mathematical reasoning problems used during training.

\begin{rolloutbox}[lvlone, label={roll:fs}]{\Lone{} on FrontierScience}
\turnbar{Problem \texttt{chemistry:72}, reference answer \texttt{-8}}
\begin{rtext}
Calculate the change in oxidation state of sulfur when potassium peroxydisulfate (K2S2O8) is reduced to potassium sulfide (K2S).
\end{rtext}
\turnbar{Root, step 1 --- delegate}
\begin{ptext}
I will have multiple subagents independently determine the oxidation state of sulfur in K2S2O8 and K2S, then compute the change when reduced, to cross-check the result.

\textbf{TASK:} Calculate the oxidation state of sulfur in potassium peroxydisulfate (K2S2O8) and in potassium sulfide (K2S, where sulfur is in the -2 oxidation state), and determine the change in oxidation state of sulfur when K2S2O8 is reduced to K2S. Show your reasoning step by step and give the numerical change.

\textbf{TASK:} Determine the oxidation number of sulfur in K2S2O8 (potassium peroxydisulfate) and in K2S (potassium sulfide), then find the change in oxidation state of sulfur per sulfur atom when K2S2O8 is reduced to K2S. Provide the calculation and the final numerical value for the change.

\textbf{TASK:} Solve this chemistry problem independently: Calculate the change in oxidation state of sulfur when potassium peroxydisulfate (K2S2O8) is reduced to potassium sulfide (K2S). Show your work clearly and state the final numerical change.
\end{ptext}
\turnbar{3 subagents run in parallel in fresh 8K windows $\rightarrow$ reports returned to the root at step 2}
\begin{rtext}
\textbf{Task 1.} Calculate the oxidation state of sulfur in potassium peroxydisulfate (K2S2O8) and in potassium sulfide (K2S, where sulfur is in the -2 oxidation state), and determine the change in oxidation state of sulfur when K2S2O8 is reduced to K2S. Show your reasoning step by step and give the numerical change.
\end{rtext}
\elide{subagent 1 reasoning: 6,245 characters omitted}
\begin{ptext}
\#\#\# RESULT 
Scope: Determined oxidation states of sulfur in K\ensuremath{_{2}}S\ensuremath{_{2}}O\ensuremath{_{8}} and K\ensuremath{_{2}}S, and the change in oxidation state when K\ensuremath{_{2}}S\ensuremath{_{2}}O\ensuremath{_{8}} is reduced to K\ensuremath{_{2}}S.

Finding: Sulfur oxidation state in K\ensuremath{_{2}}S\ensuremath{_{2}}O\ensuremath{_{8}} is +6 (due to peroxide oxygen at -1), in K\ensuremath{_{2}}S is -2; the change per sulfur atom is from +6 to -2, a decrease of 8.

Answer: \textbackslash{}boxed\{-8\}
\end{ptext}
\begin{rtext}
\textbf{Task 2.} Determine the oxidation number of sulfur in K2S2O8 (potassium peroxydisulfate) and in K2S (potassium sulfide), then find the change in oxidation state of sulfur per sulfur atom when K2S2O8 is reduced to K2S. Provide the calculation and the final numerical value for the change.
\end{rtext}
\elide{subagent 2 reasoning: 2,787 characters omitted}
\begin{ptext}
\#\#\# RESULT
Scope: Determined oxidation numbers of sulfur in K2S2O8 and K2S, and computed the change per sulfur atom when K2S2O8 is reduced to K2S.

Finding: Sulfur in K2S2O8 is +6 (due to peroxide linkage in S2O8\textasciicircum{}2-), and in K2S it is -2; the change per sulfur atom is -8.

Answer: \textbackslash{}boxed\{-8\}
\end{ptext}
\begin{rtext}
\textbf{Task 3.} Solve this chemistry problem independently: Calculate the change in oxidation state of sulfur when potassium peroxydisulfate (K2S2O8) is reduced to potassium sulfide (K2S). Show your work clearly and state the final numerical change.
\end{rtext}
\elide{subagent 3 reasoning: 2,724 characters omitted}
\begin{ptext}
\#\#\# RESULT 
Scope: Calculated the change in oxidation state of sulfur when K\ensuremath{_{2}}S\ensuremath{_{2}}O\ensuremath{_{8}} is reduced to K\ensuremath{_{2}}S, by determining the oxidation numbers of sulfur in each compound and comparing them.

Finding: In K\ensuremath{_{2}}S\ensuremath{_{2}}O\ensuremath{_{8}}, sulfur has an oxidation state of +6; in K\ensuremath{_{2}}S, sulfur has an oxidation state of -2. The change is a decrease of 8, so the magnitude of the change is 8.

Answer: \textbackslash{}boxed\{-8\}
\end{ptext}
\turnbar{Root, step 2 --- provisional answer, then delegate}
\begin{ptext}
\#\#\# ANSWER

All three subagents agree: sulfur goes from +6 in K2S2O8 to -2 in K2S, a change of -8 per sulfur atom.

\textbackslash{}boxed\{-8\}

I will verify this consensus with one more independent check to confirm the peroxide oxygen assignment and arithmetic are correct.

\textbf{TASK:} Verify the following calculation for a chemistry problem: In potassium peroxydisulfate (K2S2O8), determine the oxidation state of sulfur. The ion is S2O8\textasciicircum{}2-. Potassium is +1 each. Oxygen in peroxydisulfate is in a peroxide linkage (O-O bond), so each oxygen has oxidation state -1. Using this, calculate the oxidation state of sulfur. Then, in potassium sulfide (K2S), determine the oxidation state of sulfur. Then compute the change in oxidation state of sulfur when K2S2O8 is reduced to K2S. Confirm whether the change per sulfur atom is -8. Show all steps.
\end{ptext}
\turnbar{1 subagents run in parallel in fresh 8K windows $\rightarrow$ reports returned to the root at step 3}
\begin{rtext}
\textbf{Task 1.} Verify the following calculation for a chemistry problem: In potassium peroxydisulfate (K2S2O8), determine the oxidation state of sulfur. The ion is S2O8\textasciicircum{}2-. Potassium is +1 each. Oxygen in peroxydisulfate is in a peroxide linkage (O-O bond), so each oxygen has oxidation state -1. Using this, calculate the oxidation state of sulfur. Then, in potassium sulfide (K2S), determine the oxidation state of sulfur. Then compute the change in oxidation state of sulfur when K2S2O8 is reduced to K2S. \elide{66 more characters omitted}
\end{rtext}
\elide{subagent 1 reasoning: 1,807 characters omitted}
\begin{ptext}
\#\#\# RESULT
Scope: Verified oxidation states of sulfur in K\ensuremath{_{2}}S\ensuremath{_{2}}O\ensuremath{_{8}} (peroxydisulfate) and K\ensuremath{_{2}}S, and the change in oxidation state per sulfur atom upon reduction.

Finding: Sulfur oxidation state in S\ensuremath{_{2}}O\ensuremath{_{8}}\ensuremath{^{2}}\ensuremath{^{-}} is +6, in K\ensuremath{_{2}}S is -2, so the change per sulfur atom is -8, matching the problem's claim.

Answer: \textbackslash{}boxed\{-8\}
\end{ptext}
\turnbar{Root, step 3 --- return}
\begin{ptext}
\#\#\# RETURN

All four independent subagents confirm sulfur's oxidation state changes from +6 in K2S2O8 to -2 in K2S, a decrease of 8 per sulfur atom.

\textbackslash{}boxed\{-8\}
\end{ptext}
\end{rolloutbox}

\subsection{\Lone Rollouts under Larger Inference-time Compute Budgets}
\label{app:rollouts:additional-compute}

\learning trains with $B = 3$ subagents per delegation step, $W = 1$ context windows, and delegation depth $D = 1$. In this section, we provide example rollouts under \Lone with larger inference-time compute budgets than seen during training: 4 subagents per delegation step (Rollout~\ref{roll:f4}), 3 context windows connected through digestion (Rollout~\ref{roll:digest}), and delegation depth 2 (Rollout~\ref{roll:depth}). See their corresponding experimental results in Sec.~\ref{sec:exp:generalization}

In Rollout~\ref{roll:f4}, the model exploits the additional delegation opportunity by assigning the extra slot to a genuinely distinct task, suggesting that it has learned contextual reasoning behaviors that generalize beyond the training configuration. This rollout also provides an example of a delegation step containing \emph{mixed} search strategies: two tasks specify explicit solution techniques while two do not, combining both \texttt{Same} and \texttt{Different}.

\begin{rolloutbox}[lvlone, label={roll:f4}]{\Lone{} with 4 subagents ($B = 4$)}
\turnbar{Problem \texttt{aime\_2024:72}, reference answer \texttt{540}}
\begin{rtext}
Find the largest possible real part of \[(75+117i)z+\frac{96+144i}{z}\]where $z$ is a complex number with $|z|=4$.
\end{rtext}
\turnbar{Root, step 1 --- delegate}
\begin{ptext}
I will dispatch independent subagents to solve this complex number optimization problem using different approaches so I can cross-check their answers.

\textbf{TASK:} Solve the following competition math problem completely and show your work: Find the largest possible real part of (75+117i)z + (96+144i)/z where z is a complex number with |z|=4.

\textbf{TASK:} Solve the following competition math problem completely using the substitution z = 4e\textasciicircum{}\{i\ensuremath{\theta}\} and trigonometric manipulation: Find the largest possible real part of (75+117i)z + (96+144i)/z where z is a complex number with |z|=4.

\textbf{TASK:} Solve the following competition math problem completely by writing z = x+iy with x\textasciicircum{}2+y\textasciicircum{}2=16 and using calculus/algebraic optimization techniques: Find the largest possible real part of (75+117i)z + (96+144i)/z where z is a complex number with |z|=4.

\textbf{TASK:} Solve the following competition math problem completely, showing full derivation: Find the largest possible real part of (75+117i)z + (96+144i)/z where z is a complex number with |z|=4.
\end{ptext}
\turnbar{4 subagents run in parallel in fresh 8K windows $\rightarrow$ reports returned to the root at step 2}
\begin{rtext}
\textbf{Task 1.} Solve the following competition math problem completely and show your work: Find the largest possible real part of (75+117i)z + (96+144i)/z where z is a complex number with |z|=4.
\end{rtext}
\elide{subagent 1 reasoning: 4,278 characters omitted}
\begin{ptext}
\#\#\# RESULT
Scope: Find the maximum real part of \$ (75+117i)z + (96+144i)/z \$ given \$ |z|=4 \$.

Finding: Writing \$ z=4e\textasciicircum{}\{i\textbackslash{}theta\} \$, the expression becomes \$4[(75+117i)e\textasciicircum{}\{i\textbackslash{}theta\}+(6+9i)e\textasciicircum{}\{-i\textbackslash{}theta\}]\$, whose real part simplifies to \$4(81\textbackslash{}cos\textbackslash{}theta-108\textbackslash{}sin\textbackslash{}theta)\$, maximized by the amplitude \$ \textbackslash{}sqrt\{81\textasciicircum{}2+108\textasciicircum{}2\}=135 \$.

Answer: \textbackslash{}boxed\{540\}
\end{ptext}
\begin{rtext}
\textbf{Task 2.} Solve the following competition math problem completely using the substitution z = 4e\textasciicircum{}\{i\ensuremath{\theta}\} and trigonometric manipulation: Find the largest possible real part of (75+117i)z + (96+144i)/z where z is a complex number with |z|=4.
\end{rtext}
\elide{subagent 2 reasoning: 2,339 characters omitted}
\begin{ptext}
\#\#\# RESULT
Scope: Find the maximum real part of \$ (75+117i)z + (96+144i)/z \$ given \$ |z|=4 \$, using substitution \$ z=4e\textasciicircum{}\{i\textbackslash{}theta\} \$.

Finding: The expression simplifies to \$ 324\textbackslash{}cos\textbackslash{}theta - 432\textbackslash{}sin\textbackslash{}theta \$, whose maximum amplitude is \$ \textbackslash{}sqrt\{324\textasciicircum{}2+432\textasciicircum{}2\}=540 \$.

Answer: \textbackslash{}boxed\{540\}
\end{ptext}
\begin{rtext}
\textbf{Task 3.} Solve the following competition math problem completely by writing z = x+iy with x\textasciicircum{}2+y\textasciicircum{}2=16 and using calculus/algebraic optimization techniques: Find the largest possible real part of (75+117i)z + (96+144i)/z where z is a complex number with |z|=4.
\end{rtext}
\elide{subagent 3 reasoning: 3,413 characters omitted}
\begin{ptext}
\#\#\# RESULT
Scope: Find the maximum real part of \$ (75+117i)z + (96+144i)/z \$ given \$ |z|=4 \$, using calculus/algebraic optimization.

Finding: Expressed in polar form \$ z=4e\textasciicircum{}\{i\textbackslash{}theta\} \$, the real part simplifies to \$ 324\textbackslash{}cos\textbackslash{}theta - 432\textbackslash{}sin\textbackslash{}theta \$, whose amplitude is \$ \textbackslash{}sqrt\{324\textasciicircum{}2+432\textasciicircum{}2\}=540 \$.

Answer: \textbackslash{}boxed\{540\}
\end{ptext}
\begin{rtext}
\textbf{Task 4.} Solve the following competition math problem completely, showing full derivation: Find the largest possible real part of (75+117i)z + (96+144i)/z where z is a complex number with |z|=4.
\end{rtext}
\elide{subagent 4 reasoning: 4,127 characters omitted}
\begin{ptext}
\#\#\# RESULT
Scope: Find the maximum real part of \$ (75+117i)z + (96+144i)/z \$ given \$ |z|=4 \$, using a trigonometric parametrization and linear combination approach.

Finding: The real part simplifies to \$ 324\textbackslash{}cos\textbackslash{}theta - 432\textbackslash{}sin\textbackslash{}theta \$, whose maximum is \$ \textbackslash{}sqrt\{324\textasciicircum{}2+432\textasciicircum{}2\}=540 \$, achieved for some \$ z \$ on \$ |z|=4 \$.

Answer: \textbackslash{}boxed\{540\}
\end{ptext}
\turnbar{Root, step 2 --- provisional answer, then delegate}
\begin{ptext}
\#\#\# ANSWER
All four subagents independently derived the same simplified expression and amplitude, agreeing the maximum real part is 540.

\textbackslash{}boxed\{540\}

I will verify this consensus answer with an independent check on the algebra and the amplitude calculation to ensure no arithmetic slip occurred.

\textbf{TASK:} Verify the following algebra carefully: given z=4e\textasciicircum{}\{i\ensuremath{\theta}\}, compute the real part of (75+117i)z + (96+144i)/z step by step, and confirm whether it simplifies to 324cos\ensuremath{\theta} - 432sin\ensuremath{\theta}, and confirm the amplitude sqrt(324\textasciicircum{}2+432\textasciicircum{}2) equals 540.

\textbf{TASK:} Independently re-derive from scratch, showing full work, the maximum real part of (75+117i)z + (96+144i)/z given |z|=4, and state the final numerical answer clearly.
\end{ptext}
\turnbar{2 subagents run in parallel in fresh 8K windows $\rightarrow$ reports returned to the root at step 3}
\begin{rtext}
\textbf{Task 1.} Verify the following algebra carefully: given z=4e\textasciicircum{}\{i\ensuremath{\theta}\}, compute the real part of (75+117i)z + (96+144i)/z step by step, and confirm whether it simplifies to 324cos\ensuremath{\theta} - 432sin\ensuremath{\theta}, and confirm the amplitude sqrt(324\textasciicircum{}2+432\textasciicircum{}2) equals 540.
\end{rtext}
\elide{subagent 1 reasoning: 3,637 characters omitted}
\begin{ptext}
\#\#\# RESULT 
Scope: Verified that for \textbackslash{}( z=4e\textasciicircum{}\{i\textbackslash{}theta\} \textbackslash{}), the real part of \textbackslash{}( (75+117i)z + (96+144i)/z \textbackslash{}) simplifies to \textbackslash{}( 324\textbackslash{}cos\textbackslash{}theta - 432\textbackslash{}sin\textbackslash{}theta \textbackslash{}), and checked that its amplitude \textbackslash{}( \textbackslash{}sqrt\{324\textasciicircum{}2+432\textasciicircum{}2\} = 540 \textbackslash{}).

Finding: The algebra is correct: the real part is \textbackslash{}( 324\textbackslash{}cos\textbackslash{}theta - 432\textbackslash{}sin\textbackslash{}theta \textbackslash{}), and \textbackslash{}( \textbackslash{}sqrt\{324\textasciicircum{}2+432\textasciicircum{}2\} = \textbackslash{}sqrt\{291600\} = 540 \textbackslash{}).

Answer: \textbackslash{}boxed\{324\textbackslash{}cos\textbackslash{}theta - 432\textbackslash{}sin\textbackslash{}theta\} and amplitude \textbackslash{}(\textbackslash{}sqrt\{324\textasciicircum{}2+432\textasciicircum{}2\}=540\textbackslash{}) both confirmed.
\end{ptext}
\begin{rtext}
\textbf{Task 2.} Independently re-derive from scratch, showing full work, the maximum real part of (75+117i)z + (96+144i)/z given |z|=4, and state the final numerical answer clearly.
\end{rtext}
\elide{subagent 2 reasoning: 4,816 characters omitted}
\begin{ptext}
\#\#\# RESULT 
Scope: Maximum real part of \$(75+117i)z+(96+114i)/z\$ given \$|z|=4\$, re-derived via trigonometric expansion and amplitude maximization.

Finding: The real part simplifies to \$324\textbackslash{}cos\textbackslash{}theta - 432\textbackslash{}sin\textbackslash{}theta\$, whose maximum amplitude is \$\textbackslash{}sqrt\{324\textasciicircum{}2+432\textasciicircum{}2\}=540\$.

Answer: \$\textbackslash{}boxed\{540\}\$
\end{ptext}
\turnbar{Root, step 3 --- return}
\begin{ptext}
\#\#\# RETURN
All subagents independently confirmed the algebra: real part = 324cos\ensuremath{\theta} - 432sin\ensuremath{\theta}, amplitude = \ensuremath{\sqrt{\phantom{x}}}(324\ensuremath{^{2}}+432\ensuremath{^{2}}) = 540.

\textbackslash{}boxed\{540\}
\end{ptext}
\end{rolloutbox}

Rollout~\ref{roll:digest} illustrates reasoning with digestion across three context windows. The first window produces an incorrect answer, and the digest enables subsequent windows to continue reasoning from the accumulated progress rather than starting from scratch. In the second window, the model revisits and verifies the earlier result and corrects the answer. The third window then builds on the new digest and reaches a consistent final answer. This example demonstrates how digestion supports reasoning across multiple context windows, allowing the model to revisit and refine prior progress as additional inference-time compute becomes available.

\begin{rolloutbox}[lvldig, label={roll:digest}]{\Lone{} across three context windows ($W=3$)}
\turnbar{Problem \texttt{aime\_2026:4}, reference answer \texttt{70} --- the same problem as Rollout~\ref{roll:l1}}
\begin{rtext}
Find the number of integers less than or equal to 100 that are equal to $a+b+ab$ for some choice of distinct positive integers a and b.
\end{rtext}
\turnbar{Window 1 --- starts with no digest; final answer 71, incorrect}
\emph{\scriptsize Root's delegation at step 1:}
\begin{ptext}
I will dispatch independent solvers to tackle this counting problem from scratch, using different approaches to cross-check the answer.

\textbf{TASK:} Find the number of integers less than or equal to 100 that can be expressed as a+b+ab for some choice of distinct positive integers a and b. Show your work step by step and give the final count.

\textbf{TASK:} Solve this competition math problem: Find the number of integers less than or equal to 100 that are equal to a+b+ab for some choice of distinct positive integers a and b. Provide full reasoning and a final numerical answer.

\textbf{TASK:} Solve the following number theory problem completely and give a final count: Find the number of integers less than or equal to 100 that are equal to a+b+ab for some choice of distinct positive integers a and b. Show all steps clearly.
\end{ptext}
\elide{2 further root steps and 6 subagents omitted}
\emph{\scriptsize Root's returned answer:}
\begin{ptext}
\#\#\# RETURN
Three independent careful analyses converged on 71, with two explicitly counting composites in [6,101] excluding prime squares (9,25,49), confirming the earlier two agreeing at 70 were missing these boundary/small cases.

\textbackslash{}boxed\{71\}
\end{ptext}
\emph{\scriptsize Digest generated for the next window (69 generated tokens):}
\begin{ptext}
\#\#\# RETURN

Three independent careful analyses now agree on 71, confirming the earlier three subagents (two at 70, one at 71) were resolved by careful detail: count composites M in [6,101] excluding prime squares (9,25,49).

\textbackslash{}boxed\{71\}
\end{ptext}
\turnbar{Window 2 --- starts with the digest of Window 1; final answer 70, correct}
\begin{ptext}
\#\#\# PRIOR WORK SUMMARY
\#\#\# RETURN

Three independent careful analyses now agree on 71, confirming the earlier three subagents (two at 70, one at 71) were resolved by careful detail: count composites M in [6,101] excluding prime squares (9,25,49).

\textbackslash{}boxed\{71\}
\end{ptext}
\emph{\scriptsize Root's delegation at step 1:}
\begin{ptext}
I want independent approaches to confirm the count of integers \ensuremath{\leq}100 expressible as a+b+ab for distinct positive integers a,b, plus a clean derivation to cross-check the prior answer of 71.

\textbf{TASK:} Solve the following problem completely and show your reasoning: Find the number of integers less than or equal to 100 that are equal to a+b+ab for some choice of distinct positive integers a and b. Provide the final count and explain the key steps of your derivation.

\textbf{TASK:} Solve the following problem completely using a substitution/algebraic approach: Find the number of integers less than or equal to 100 that are equal to a+b+ab for some choice of distinct positive integers a and b. Show your work and give the final count.

\textbf{TASK:} Solve the following problem completely and independently, showing all steps: Find the number of integers less than or equal to 100 that are equal to a+b+ab for some choice of distinct positive integers a and b. Provide the final numerical answer.
\end{ptext}
\elide{2 further root steps and 5 subagents omitted}
\emph{\scriptsize Root's returned answer:}
\begin{ptext}
\#\#\# RETURN

Four independent subagents (three solving directly, one verifying via two methods) consistently derived 70 using the substitution n+1=(a+1)(b+1), confirming the count of composite m in [4,101] excluding prime squares (74-4=70).

\textbackslash{}boxed\{70\}
\end{ptext}
\emph{\scriptsize Digest generated for the next window (53 generated tokens):}
\begin{ptext}
\#\#\# RETURN

Four independent subagents confirm via substitution and two counting methods that the count equals composite numbers in [4,101] excluding prime squares (4,9,25,49), giving 70.

\textbackslash{}boxed\{70\}
\end{ptext}
\turnbar{Window 3 --- starts with the digest of Window 2; final answer 70, correct}
\begin{ptext}
\#\#\# PRIOR WORK SUMMARY
\#\#\# RETURN

Four independent subagents confirm via substitution and two counting methods that the count equals composite numbers in [4,101] excluding prime squares (4,9,25,49), giving 70.

\textbackslash{}boxed\{70\}
\end{ptext}
\emph{\scriptsize Root's delegation at step 1:}
\begin{ptext}
I will dispatch independent subagents to solve the full problem and cross-check the structure of the solution to confirm the count of 70.

\textbf{TASK:} Solve this competition math problem completely and show your reasoning: Find the number of integers less than or equal to 100 that are equal to a+b+ab for some choice of distinct positive integers a and b.

\textbf{TASK:} Solve this competition math problem completely and show your reasoning: Find the number of integers less than or equal to 100 that are equal to a+b+ab for some choice of distinct positive integers a and b. Provide the final count as a single number and explain the key steps of your derivation.

\textbf{TASK:} Solve this competition math problem completely and show your reasoning, then give the final count as a single number: Find the number of integers less than or equal to 100 that are equal to a+b+ab for some choice of distinct positive integers a and b.
\end{ptext}
\elide{2 further root steps and 6 subagents omitted}
\emph{\scriptsize Root's returned answer:}
\begin{ptext}
\#\#\# RETURN

Two independent subagents (out of three) carefully derived 70 via the composite-counting method, matching the prior-work summary's claim; the dissenters (71) used a flawed "75 composites" count rather than the standard 74 composites in [4,101].

\textbackslash{}boxed\{70\}
\end{ptext}
\end{rolloutbox}

In Rollout~\ref{roll:depth}, subagents recursively delegate tasks to further subagents by invoking the same harness interface, illustrating extrapolation beyond the delegation depth seen during training. Despite only observing depth-1 delegation during training, the model continues to apply delegation when additional depth is enabled.

\begin{rolloutbox}[lvlsub, label={roll:depth}]{\Lone{} at delegation depth 2 ($D = 2$)}
\turnbar{Problem \texttt{aime\_2024:79}, reference answer \texttt{321}}
\begin{rtext}
Let $\omega\neq 1$ be a 13th root of unity. Find the remainder when
\[\prod_{k=0}^{12}(2-2\omega^k+\omega^{2k})\]
is divided by 1000.
\end{rtext}
\turnbar{Depth 1 --- the root delegates}
\begin{ptext}
I will dispatch multiple subagents to independently solve this competition math problem so I can cross-check their approaches and answers.

\textbf{TASK:} Let \ensuremath{\omega} \ensuremath{\neq} 1 be a 13th root of unity. Find the remainder when \ensuremath{\prod}\_\{k=0\}\textasciicircum{}\{12\}(2-2\ensuremath{\omega}\textasciicircum{}k+\ensuremath{\omega}\textasciicircum{}\{2k\}) is divided by 1000. Show full steps and give a numeric remainder.

\textbf{TASK:} Let \ensuremath{\omega} \ensuremath{\neq} 1 be a primitive 13th root of unity. Find the remainder when \ensuremath{\prod}\_\{k=0\}\textasciicircum{}\{12\}(2-2\ensuremath{\omega}\textasciicircum{}k+\ensuremath{\omega}\textasciicircum{}\{2k\}) is divided by 1000. Provide a full derivation and the final numeric answer.

\textbf{TASK:} Solve this competition math problem completely and show your work: Let \ensuremath{\omega} \ensuremath{\neq} 1 be a 13th root of unity. Find the remainder when \ensuremath{\prod}\_\{k=0\}\textasciicircum{}\{12\}(2-2\ensuremath{\omega}\textasciicircum{}k+\ensuremath{\omega}\textasciicircum{}\{2k\}) is divided by 1000.
\end{ptext}
\turnbar{Depth 2 --- task 1's subagent receives that task as its own \texttt{\#\#\# PROBLEM}, under the same \Lone harness, and it delegates again}
\begin{ptext}
I will dispatch independent subagents to solve this problem using different approaches so I can cross-check their answers.

\textbf{TASK:} Let \ensuremath{\omega} \ensuremath{\neq} 1 be a primitive 13th root of unity. Find the remainder when \ensuremath{\prod}\_\{k=0\}\textasciicircum{}\{12\}(2-2\ensuremath{\omega}\textasciicircum{}k+\ensuremath{\omega}\textasciicircum{}\{2k\}) is divided by 1000. Show full steps and give a numeric remainder.

\textbf{TASK:} Let \ensuremath{\omega} \ensuremath{\neq} 1 be a 13th root of unity (primitive, since 13 is prime and \ensuremath{\omega} \ensuremath{\neq} 1 implies it's primitive). Find the remainder when \ensuremath{\prod}\_\{k=0\}\textasciicircum{}\{12\}(2-2\ensuremath{\omega}\textasciicircum{}k+\ensuremath{\omega}\textasciicircum{}\{2k\}) is divided by 1000. Show full steps and give a numeric remainder.

\textbf{TASK:} Solve this competition math problem completely and show all steps: Let \ensuremath{\omega} \ensuremath{\neq} 1 be a 13th root of unity. Find the remainder when \ensuremath{\prod}\_\{k=0\}\textasciicircum{}\{12\}(2-2\ensuremath{\omega}\textasciicircum{}k+\ensuremath{\omega}\textasciicircum{}\{2k\}) is divided by 1000.
\end{ptext}
\turnbar{Depth 2 --- subagents reason inline}
\begin{rtext}
\textbf{Task 1.} Let \ensuremath{\omega} \ensuremath{\neq} 1 be a primitive 13th root of unity. Find the remainder when \ensuremath{\prod}\_\{k=0\}\textasciicircum{}\{12\}(2-2\ensuremath{\omega}\textasciicircum{}k+\ensuremath{\omega}\textasciicircum{}\{2k\}) is divided by 1000. Show full steps and give a numeric remainder.
\end{rtext}
\begin{ptext}
\#\#\# RESULT
Scope: The problem asks for the remainder modulo 1000 of the product \ensuremath{\prod}\_\{k=0\}\textasciicircum{}\{12\}(2-2\ensuremath{\omega}\textasciicircum{}k+\ensuremath{\omega}\textasciicircum{}\{2k\}) where \ensuremath{\omega} is a primitive 13th root of unity, \ensuremath{\omega}\ensuremath{\neq}1.

Finding: The product equals 8321, derived via resultant or symmetric function methods using roots of x\textasciicircum{}2-2x+2, confirmed by two independent approaches.

Answer: \textbackslash{}boxed\{321\}
\end{ptext}
\begin{rtext}
\textbf{Task 2.} Let \ensuremath{\omega} \ensuremath{\neq} 1 be a 13th root of unity (primitive, since 13 is prime and \ensuremath{\omega} \ensuremath{\neq} 1 implies it's primitive). Find the remainder when \ensuremath{\prod}\_\{k=0\}\textasciicircum{}\{12\}(2-2\ensuremath{\omega}\textasciicircum{}k+\ensuremath{\omega}\textasciicircum{}\{2k\}) is divided by 1000. Show full steps and give a numeric remainder.
\end{rtext}
\begin{ptext}
\#\#\# RESULT
Scope: Compute \ensuremath{\prod}\_\{k=0\}\textasciicircum{}\{12\}(2-2\ensuremath{\omega}\textasciicircum{}k+\ensuremath{\omega}\textasciicircum{}\{2k\}) for primitive 13th root of unity \ensuremath{\omega}, then find remainder mod 1000.

Finding: The product equals (\ensuremath{\forall}\textasciicircum{}13-1)(\ensuremath{\beta}\textasciicircum{}13-1) for roots \ensuremath{\alpha}=1+i, \ensuremath{\beta}=1-i of x\textasciicircum{}2-2x+2=0, computed via polar forms to be 8321.

Answer: \textbackslash{}boxed\{321\}
\end{ptext}
\elide{3 further depth-2 tasks in this branch, and 3 sibling depth-1 branches, omitted}
\turnbar{Depth 1 --- the branch reports one compact result upward}
\begin{ptext}
\#\#\# RESULT
Scope: Verified that \ensuremath{\prod}\_\{k=0\}\textasciicircum{}\{12\}(2-2\ensuremath{\omega}\textasciicircum{}k+\ensuremath{\omega}\textasciicircum{}\{2k\}) = (\ensuremath{\alpha}\textasciicircum{}13-1)(\ensuremath{\beta}\textasciicircum{}13-1) where \ensuremath{\alpha}=1+i, \ensuremath{\beta}=1-i, computed via polar form to be 8321, and 8321 mod 1000 = 321.

Finding: All independent approaches agree the product equals 8321, confirmed by two distinct methods (resultant/symmetric functions and polar form).

Answer: \textbackslash{}boxed\{321\}
\end{ptext}
\turnbar{Depth 0 --- return}
\begin{ptext}
\#\#\# RETURN

Four independent subagents confirm the product equals 8321 via multiple methods, giving remainder 8321 mod 1000 = 321.

\textbackslash{}boxed\{321\}
\end{ptext}
\end{rolloutbox}

\end{document}